\documentclass{article}

    \PassOptionsToPackage{numbers, compress}{natbib}

\usepackage[preprint]{neurips_2021}

\usepackage[utf8]{inputenc} 
\usepackage[T1]{fontenc}    
\usepackage{hyperref}       
\usepackage{url}            
\usepackage{booktabs}       
\usepackage{amsfonts}       
\usepackage{nicefrac}       
\usepackage{microtype}      
\usepackage{xcolor}         

\usepackage{microtype}
\usepackage{graphicx}
\usepackage{subfigure}
\usepackage{booktabs} 
\usepackage{outlines}
\usepackage{babel,blindtext}
\usepackage{url}
\usepackage{tabulary}
\usepackage{tabularx}
\usepackage[normalem]{ulem}
\usepackage{enumitem}
\usepackage{multirow}
\usepackage{array}
\usepackage{makecell}
\usepackage{amsmath,amsfonts,amsthm,amssymb,amsopn,bm}
\usepackage{color,soul}
\usepackage{verbatim}
\usepackage{pifont}
\usepackage{footnote}
\usepackage{wrapfig}

\newcommand\ignore[1]{}

\DeclareMathOperator*{\argmin}{arg\,min}

\title{True Few-Shot Learning with Language Models}

\author{%
  Ethan Perez$^1$, Douwe Kiela$^2$, Kyunghyun Cho$^{1 3}$\\
  $^1$New York University, $^2$Facebook AI Research,\\
  $^3$CIFAR Fellow in Learning in Machines \& Brains\\
  \texttt{perez@nyu.edu}
}

\begin{document}

\maketitle

\begin{abstract}
Pretrained language models (LMs) perform well on many tasks even when learning from a few examples, but prior work uses many held-out examples to tune various aspects of learning, such as hyperparameters, training objectives, and natural language templates (``prompts''). Here, we evaluate the few-shot ability of LMs when such held-out examples are unavailable, a setting we call \textit{true few-shot learning}. We test two model selection criteria, cross-validation and minimum description length, for choosing LM prompts and hyperparameters in the true few-shot setting. On average, both marginally outperform random selection and greatly underperform selection based on held-out examples. Moreover, selection criteria often prefer models that perform significantly worse than randomly-selected ones. We find similar results even when taking into account our uncertainty in a model's true performance during selection, as well as when varying the amount of computation and number of examples used for selection. Overall, our findings suggest that prior work significantly overestimated the true few-shot ability of LMs given the difficulty of few-shot model selection.
\end{abstract}

\section{Introduction}
\label{sec:Introduction}

Major progress in language model (LM) pretraining has led to the idea that LMs can learn a new task using a small number of examples only, i.e., few-shot learning~\citep{radford2019language,brown2020language,schick2020exploiting}.
Few-shot learning overcomes many challenges with data-rich supervised learning: collecting labeled data is expensive, often requires experts, and scales poorly with the number of tasks.
However, the few-shot performance of LMs is very sensitive to the textual task description~\citep[``prompt'';][]{schick2020exploiting,jiang-etal-2020-know,gao2020making,zhao2021calibrate}, order of training examples~\citep{zhao2021calibrate,lu2021fantastically,liu2021what}, decoding strategy~\citep{schick2020small,perez2021rissanen}, and other hyperparameters~\citep{schick2020exploiting,gao2020making,schick2020small,schick2020few,tam2021improving}, as well as the learning algorithm itself~\citep{schick2020exploiting,tam2021improving}.
Thus, effective model selection is crucial for obtaining good few-shot performance.

There are issues with how recent work approaches model selection in few-shot learning, however.
Prior work uses large train or held-out sets with many examples to choose prompts~\citep{brown2020language,tam2021improving,radford2021learning} and hyperparameters~\citep{tam2021improving}.
Other work claims to use no validation set for hyperparameter selection~\citep{schick2020exploiting,schick2020few,wang2021entailment} but does not describe how they design other aspects of their learning algorithm (e.g., training objectives).
It is unlikely that no validation examples were used, given the sophisticated nature of the proposed algorithms.
In this work, we examine if prior few-shot learning methods still perform well when using only the provided examples for model selection, a setting we term \textit{true few-shot learning}.

We find that true few-shot model selection yields prompts that marginally outperform random selection and greatly underperform selection based on held-out examples.
Our result shows that prior work may have greatly overestimated the few-shot ability of LMs.
In other words, one reason that prompts are so effective~\citep[``worth many examples'';][]{scao2021data} is that they are often tuned using many examples.
We evaluate two standard model selection criteria -- cross-validation (CV) and minimum description length (MDL) -- finding that both obtain only limited improvements over random selection and perform much worse than selection using held-out examples.
For prompt selection, our observation holds for 9 LMs ranging over 3 orders of magnitude in size~\citep{radford2019language,brown2020language,sanh2019distilbert} on 3 classification tasks and 41 tasks in the LAMA benchmark~\citep{petroni-etal-2019-language}.
For choosing hyperparameters, true few-shot selection causes performance to drop by 2-10\% across 8 tasks for ADAPET~\citep{tam2021improving}, a state-of-the-art few-shot method.
Furthermore, true few-shot model selection has high variance in performance; selected models often do much worse than randomly-chosen ones.
We find similar results when varying the number of examples used, amount of computation, and conservativeness of our selection criterion.
Altogether, our results suggest that model selection is a fundamental roadblock to true few-shot learning.

\section{Can We Do Model Selection in Few-Shot Learning?}
\label{sec:Background}


Prior work uses the phrase ``few-shot learning'' in multiple senses, raising questions about what it means to do few-shot learning.
We categorize few-shot learning into three distinct settings, each of which assumes access to different data.
Here, we formally disambiguate between these settings to help future work avoid inadvertently comparing few-shot methods that operate in different settings.

Consider the supervised learning scenario where we have a dataset of inputs $x_{1:N}$ and labels $y_{1:N}$, sampled from a distribution over datasets $D$.
We aim to determine the learning algorithm $\mathcal{A}^* \in \mathcal{A}_1, \dots, \mathcal{A}_A$ with the smallest generalization loss $\mathcal{L}$ at predicting $y$ given $x$ on unseen validation examples $D_{\text{val}} \sim D$ after learning on training examples $D_{\text{train}} \sim D$.
We say that an algorithm $\mathcal{A}(D_{\text{train}}, R)$ maps a training dataset $D_{\text{train}}$ and various random factors $R$ that influence training to a function that predicts $y$ given $x$.
$\mathcal{A}$ specifies, e.g., the model architecture, hyperparameters, and prompt.
$R$ includes random factors that impact the results of a learning algorithm, such as parameter initialization and the order of training examples for online learning algorithms like stochastic gradient descent.
We say that $\mathcal{A}$ obtains a generalization loss $\mathcal{L}(\mathcal{A}(D_{\text{train}}, R), D_{\text{val}})$ on a given validation set $D_{\text{val}}$.
We aim to find the $\mathcal{A}^*$ that minimizes the expected loss across training and validation sets:
\begin{align*}
    \text{EL}(\mathcal{A}, R) = \mathbb{E}_{D_{\text{train}}, D_{\text{val}}} \bigg[\mathcal{L}\Big(\mathcal{A}(D_{\text{train}}, R); D_{\text{val}}\Big)\bigg]
\end{align*}

In \textit{data-rich supervised learning}, $\text{EL}(\mathcal{A}, R)$ is usually evaluated with a single train-validation split $(D_{\text{train}}, D_{\text{val}})$.
Since large $D_{\text{train}}$ and $D_{\text{val}}$ are not always available, the traditional few-shot setting evaluates $\text{EL}(\mathcal{A}, R)$ with many small $(D_{\text{train}}, D_{\text{val}})$ drawn from many, distinct distributions $D$~\citep[see, e.g., work in meta-learning][]{vinyals2016matching,snell2017prototypical,ravi2017optimization,li2017learning}.
Each distribution $D$ is sampled from $D^*$, a distribution over distributions (e.g., of similar tasks), so we call this setting \textit{multi-distribution few-shot learning}.

Recent work does not assume access to data from other distributions, performing few-shot learning using only a few examples from a single distribution to update a pretrained LM~\citep{brown2020language,tam2021improving}.
These papers use a large validation set $D_{\text{val}}$ to tune the learning algorithm $\mathcal{A}$, a setting we term \textit{tuned few-shot learning}.
For example, Brown et al.~\citep{brown2020language} try prompts with different phrasings and numbers of training examples to improve the validation accuracy of GPT-3.
Tam et al.~\citep{tam2021improving} choose the early stopping iteration, prompt, and other model-specific hyperparameters based on validation performance.
Tuned few-shot learning relies on many labeled examples, so we argue that tuned few-shot learning does not qualify as few-shot learning.
If many validation examples are available, they could be incorporated into the training set and trained on using data-rich supervised learning.
Tuned few-shot learning algorithms should be compared against data-rich supervised learning algorithms that use the same amount of total data $|D_{\text{train}}|+|D_{\text{val}}|$.

In this work, we evaluate the success of tuned few-shot learning methods when no large $D_{\text{val}}$ is available, a setting we term \textit{true few-shot learning}.
Formally, we aim to choose a learning algorithm $\mathcal{A}$ with low expected loss $\text{EL}(\mathcal{A}, R)$ using only a small training set $D_{\text{train}}$ drawn from a single distribution.
Here, we must choose $\mathcal{A}$ by approximating $\text{EL}(\mathcal{A}, R)$, e.g., using cross-validation.
Several papers claim to circumvent the need to estimate $\text{EL}(\mathcal{A}, R)$ by choosing hyperparameters based on an educated guess~\citep{schick2020exploiting,schick2020small,wang2021entailment}.
However, the proposed learning algorithms themselves are quite sophisticated, and it is unclear how they were designed if not by using validation performance.
Other work chooses the learning algorithm and hyperparameters using one or multiple other datasets before evaluating on the target dataset~\citep{gao2020making,schick2020few}.
Such approaches fall under \textit{multi-distribution few-shot learning} and cannot be directly compared to methods that attempt to perform true few-shot learning, even though prior work has made such comparisons~\citep{wang2021entailment}.

In what follows, we describe two model selection criteria -- cross-validation and minimum description length -- which we use to evaluate tuned few-shot methods in the true few-shot setting.

\subsection{Cross-validation}
\label{ssec:Cross Validation (CV)}

Cross-Validation (CV)~\citep{allend1974relationship,stone1974cross,geisser1975predictive} is one of the most widely used methods for estimating generalization loss~\citep{hastie2001statistical}.
CV has also been used in prior work on multi-distribution few-shot learning~\citep{finn2017model,rajeswaran2019meta}.
CV randomly partitions $D_{\text{train}}$ into $K$ equally-sized folds $F(D_{\text{train}})_1, \dots, F(D_{\text{train}})_K$ and evaluates the average loss on a validation fold $F(D_{\text{train}})_k$ when training on the remaining data $F(D_{\text{train}})_{\neg k}$:
\begin{align*}
    \text{CV}(\mathcal{A},R,F) = \mathbb{E}_{k \sim \text{Unif}(1, K)} \bigg[ \mathcal{L} \Big( \mathcal{A}(F(D_{\text{train}})_{\neg k}, R); F(D_{\text{train}})_{k} \Big) \bigg]
\end{align*}

In this way, CV forms $K$ train-validation splits out of the pool of labeled examples.
CV with one example per fold ($K=N$ folds) is commonly referred to as leave-one-out CV (LOOCV).

\subsection{Minimum description length}
\label{ssec:Minimum Description Length (MDL)}

We may also form train-validation splits in a different manner than CV, drawing inspiration from work on the Minimum Description Length (MDL) principle~\citep{rissanen1978modeling}.
MDL can be estimated by evaluating the average loss on a fold $F(D)_k$ when training on the previous folds $F(D)_{1:k-1}$:
\begin{align*}
    \text{MDL}(\mathcal{A},R,F) = \mathbb{E}_{k \sim \text{Unif}(1, K)} \bigg[ \mathcal{L} \Big( \mathcal{A}(F(D_{\text{train}})_{1:k-1}, R); F(D_{\text{train}})_{k} \Big) \bigg]
\end{align*}

This procedure is referred to as ``online coding''~\citep{rissanen1984universal,dawid1984present}, as it evaluates the generalization loss of the algorithm as it learns ``online'' from more and more data.\footnote{Online coding formally computes a sum over $\mathcal{L}(.)$ rather than the expectation, which differs by a constant factor. The two are equivalent for our purposes (ranking $\mathcal{A}$).
}
There are other ways to evaluate MDL~\citep[see][for an overview]{grunwald2004tutorial}. We use online coding as it has been shown to be an effective way to estimate MDL, especially for deep learning methods~\citep{blier2018description}.

MDL measures generalization because it evaluates how much a learning algorithm compresses the labels $y_{1:N}$ given the inputs $x_{1:N}$, and because better compression implies better generalization~\citep{blumer1987occam}.
Recent work has used MDL to determine which learning algorithms are most effective at explaining the given data~\citep[Rissanen Data Analysis;][]{perez2021rissanen,sinha2021masked}.

\subsection{Variance matters}
\label{ssec:Variance Matters}

We evaluate the generalization loss of the algorithm chosen by CV (likewise for MDL):
\begin{align*}
    \mathcal{L}(\mathcal{A}_{\text{CV}}(D_{\text{train}}, R), D_{\text{val}}), && \text{where } \mathcal{A}_{\text{CV}} = \argmin_{\mathcal{A}} \mathbb{E}_{R,F}[\text{CV}(\mathcal{A}, R, F)].
\end{align*}

The above loss should be low in expectation, across different datasets $D_{\text{train}} \sim D$, $D_{\text{val}} \sim D$, and random factors $R, F$: $\mathbb{E}_{D_{\text{train}}, D_{\text{val}}, R, F} [\mathcal{L}(\mathcal{A}_{\text{CV}}(D_{\text{train}}, R), D_{\text{val}})]$.
The loss should also be low in variance: $\mathbb{V}_{D_{\text{train}}, D_{\text{val}}, R, F} [\mathcal{L}(\mathcal{A}_{\text{CV}}(D_{\text{train}}, R), D_{\text{val}})]$.
Low variance implies that CV/MDL \textit{reliably} choose an algorithm that generalizes to $D_{\text{val}}$ when trained with a given $D_{\text{train}}$ and random factors $R,F$.
Reliability is important for many practical or commercial applications where worst-case performance is important, such as image recognition~\citep{phillips2000feret,buolamwini2018gender}, dialogue systems~\citep{henderson2018ethical,khatri2018advancing}, and robotics~\citep{garcia2015comprehensive,amodei2016concrete}.

We also experiment with explicitly taking into account an algorithm's variance during model selection, choosing $\mathcal{A}_{\text{CV}}$ to minimize a conservative estimate of CV, $\text{CV}_{\alpha}(\mathcal{A})$, chosen such that the probability $\text{Pr}_{R,F}[\text{CV}(\mathcal{A}, R, F) < \text{CV}_{\alpha}(\mathcal{A})]$ is high:
\begin{align*}
    \text{CV}_{\alpha}(\mathcal{A}) = \mathbb{E}_{R,F}[ \text{CV}(\mathcal{A}, R, F) ] + \alpha \sqrt{\mathbb{V}_{R,F}[ \text{CV}(\mathcal{A}, R, F) ]}
\end{align*}
where $\alpha$ is a hyperparameter set based on the desired probability.
In particular, if $\text{CV}(\mathcal{A}, R, F)$ follows a normal distribution $\mathcal{N}$ when sampling $R,F$, then $\text{CV}(\mathcal{A}, R, F) \leq \text{CV}_{\alpha}(\mathcal{A})$ with probability $\int_{-\infty}^{\alpha} \mathcal{N}(\mu=0, \sigma=1)$ for a given $R,F$.
$\text{CV}_{\alpha}(\mathcal{A})$ resembles the Watanabe Akaike Information Criterion~\citep{watanabe2010asymptotic}, which estimates the generalization of a model trained with $\mathcal{A}$ using the expected loss from a model trained with $\mathcal{A}$ plus the variance in training loss across models trained with $\mathcal{A}$.

\subsection{Other model selection criteria}
\label{ssec:Other Model Selection Criteria}
Prior work has developed other model selection criteria such as the Akaike Information Criterion~\citep[AIC;][]{akaike1974new}, Watanabe-Akaike Information Criterion~\citep[WAIC;][]{watanabe2010asymptotic}, and Mallows' $C_p$~\citep{mallows1973some}.
These methods often rely on assumptions or quantities that are not available in the context of deep learning (AIC, Mallows' $C_p$) or are approximations of LOOCV (WAIC).
Since state-of-the-art few-shot learning methods tend to be based on deep learning, we focus on CV and MDL as our model selection criteria.
In Appendix \S\ref{sec:True Few-Shot Prompt Selection with Other Generalization Criteria}, we also test several other criteria that are applicable to deep learning methods.

Selection criteria can be optimized automatically, e.g. with bayesian optimization~\citep[][]{hutter2011sequential,bergstra2011algorithms,snoek2012practical}, evolutionary methods~\citep{bergstra2011algorithms,miikkulainen2019evolving,real2019regularized}, reinforcement learning~\citep{zoph2017neural}, or gradient descent~\citep{larsen1996design,bengio2000gradient,chapelle2004choosing,liu2018darts}.
Such methods aim to match the performance of exhaustive search, the optimal approach (used in our work).

\section{True Few-Shot Prompt Selection}
\label{sec:True Few-Shot Prompt Selection}

Recent work on LMs performs few-shot learning by providing training examples as input in the form of a natural language ``prompt''~\citep{brown2020language,schick2020exploiting,schick2020small}.
For example, for a question-answering task, Brown et al.~\cite{brown2020language} prepend input examples with ``READING COMPREHENSION ANSWER KEY'' before providing them to GPT-3 (see Appendix Table~\ref{tab:prompts} for more examples).
They then have the LM complete the remaining words in the prompt, conditioning on earlier words (including various input examples), following the LM's pretraining objective (next word prediction).
No parameter updates are involved.
It is not obvious \emph{a priori} which prompts will generalize well for a given LM, and there is high variance in how well different prompts generalize~\citep{schick2020exploiting,zhao2021calibrate}, even between prompts with minor differences~\citep[e.g., one comma;][]{gao2020making}.
Thus, it is important to choose prompts using a limited number of labeled examples to achieve true few-shot learning.

\subsection{Experimental setup}
\label{ssec:Experimental Setup}

In what follows, we test on LAMA~\citep{petroni-etal-2019-language}, a benchmark for retrieving facts with LMs, for which prior work has developed many strategies for designing prompts~\citep{jiang-etal-2020-know,shin-etal-2020-autoprompt,liu2021gpt,zhong2021factual}.
LAMA evaluates the accuracy of LMs at choosing the correct target object for various (\texttt{subject}, \texttt{relation}, \texttt{object}) triples present in knowledge bases, such as (\texttt{Dante}, \texttt{born-in}, \texttt{Florence}).
We use the ``TREx'' split, which consists of 41 relations (up to 1k examples each).
Petroni et al.~\cite{petroni-etal-2019-language} design a prompt for each relation, which an LM completes to predict an answer (e.g., ``The birthplace of Dante was \_'').
Some relations have multiple valid target entities, so LAMA evaluates how often one of the true answers matches the top-predicted token (out of 20k candidates).
We only use examples from the LAMA-UnHelpfulNames subset~\citep[LAMA-UHN;][]{poerner-etal-2020-e} which filters out easy-to-guess examples (e.g., ``The Apple Watch was created by \_'' with the answer \textit{Apple}).
We test the 5-shot accuracy of 9 popular LMs of various sizes: GPT-3~\citep[175B, 13B, 6.7B, 2.7B parameter models;][]{brown2020language}, GPT-2~\citep[1.5B, 782M, 345M, 117M models;][]{brown2020language}, and DistilGPT-2~\citep{sanh2019distilbert}, a distilled, 82M parameter version of GPT-2 117M.\footnote{We use OpenAI's API for GPT-3 (\url{https://beta.openai.com/}) and HuggingFace Transformers~\citep{wolf-etal-2020-transformers} via PyTorch~\citep{paszke2019pytorch} for GPT-2 and DistilGPT-2. OpenAI does not disclose the sizes of their API-provided models, so we follow prior work~\citep{zhao2021calibrate,lu2021fantastically} and assume that the four API models are the four largest ones from Brown et al.~\citep{brown2020language}. We plan to update our paper should OpenAI release model details.}

\paragraph{Prompts}
To form our set of candidate prompts $\mathcal{A}_1, \dots, \mathcal{A}_A$, we rely on LAMA as well as the Language model Prompt And Query Archive~\citep[LPAQA;][]{jiang-etal-2020-know}.
For each relation, we use the manually-written prompt from LAMA, as well as LPAQA prompts formed by (1) paraphrasing the manual prompt using back-translation (2) mining from Wikipedia, and (3) paraphrasing the top mined prompt.
For each relation, we use up to 16 prompts with a mean of 12 prompts (see Appendix \S\ref{ssec:LAMA} for more details on the prompts we use).

\paragraph{Computing CV and MDL}
As the loss function $\mathcal{L}$, we use the negative log-likelihood (NLL) of the label given the input over all evaluation examples $\sum_{(x, y)} -\log p(y|x)$.
We use NLL following prior work in MDL~\citep[][]{blier2018description,voita-titov-2020-information,perez2021rissanen}, to retain MDL's property as a measure of label compression.
For CV, NLL avoids ties between different prompts that would arise from using accuracy in the context of such limited data (e.g., 5 examples).
For all prompt experiments, we use $K=N$ folds (where $N$ is the number of training examples) for both MDL and CV (here, LOOCV).
Here, $N$-fold CV requires $N$ forward passes to evaluate the loss on each of the $N$ examples when conditioning on the $N-1$ other examples.
$N$-fold MDL can be computed using a single LM forward pass to compute the loss on each example conditioned on the previous examples.
This feature makes MDL more computationally efficient than CV, and enables us to compute more estimates of MDL given a fixed compute budget.

\paragraph{Marginalizing out example order}
The order of training examples impacts the generalization of LMs~\citep{lu2021fantastically}, so we treat order as a random factor $R$ that we marginalize over to evaluate the generalization of a prompt $\mathcal{A}$.
We compute the exact $\mathbb{E}_{R,F}[\text{CV}(\mathcal{A}, R, F)]$ and $\mathbb{E}_{R,F}[\text{MDL}(\mathcal{A}, R, F)]$ by averaging over all $N!$ training example orders.
We use $N=5$ examples to limit $N!$.
We estimate the average test accuracy on $N!=120$ examples in LAMA, excluding the training examples, by evaluating on one test example per permutation of training examples.
We compute CV, MDL, and test accuracy with $N!=120$ forward passes in total by appending a test example to each permutation of training examples, and we compute all selection criteria using the same set of $N!=120$ forward passes to maximize comparability across different methods.
We show the test accuracy from CV/MDL-chosen prompts, averaged over all relations.
For comparison, we show the test accuracy of always choosing (1) the best prompt, chosen using held-out accuracy as in prior work, (2) the worst prompt, as a lower bound, and (3) random prompts (we show the mean accuracy over all prompts).

\subsection{How well does prompt selection do in true few-shot learning?}
\label{ssec:How well does true few-shot prompt selection do?}
\begin{figure*}[t]
	\centering
    \includegraphics[scale=.48]{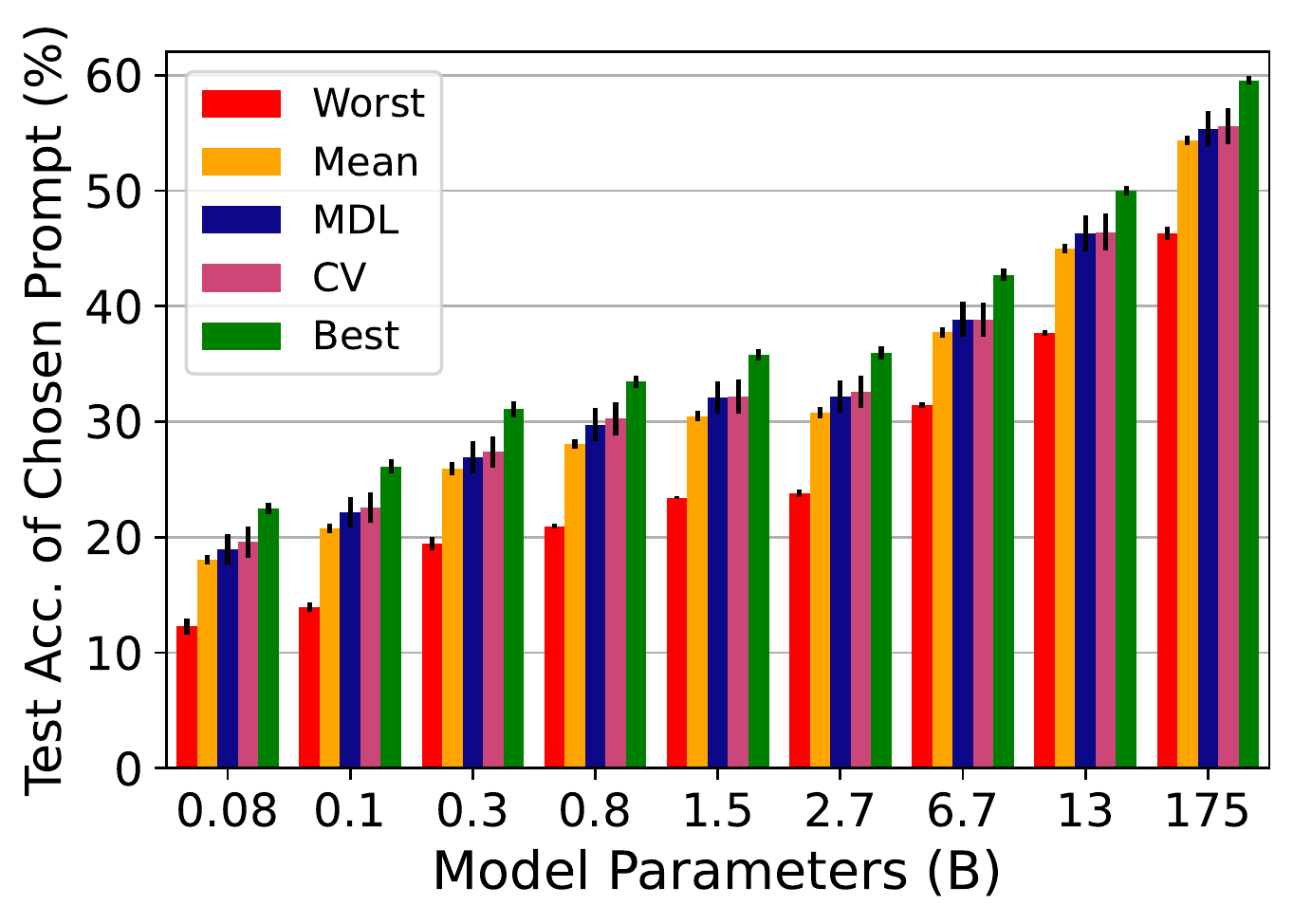}
    \includegraphics[scale=.48]{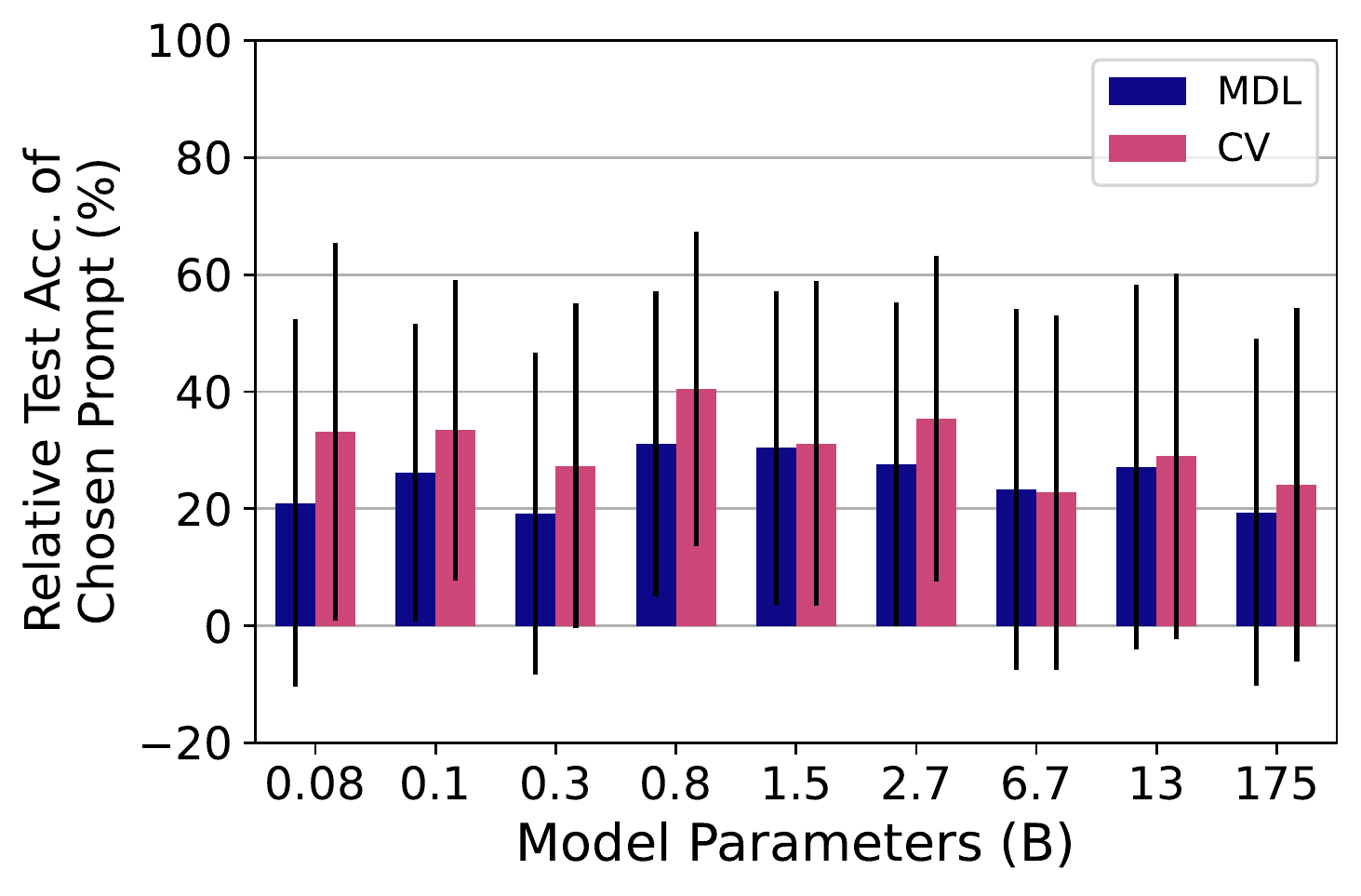}
    \caption{\textbf{Left}: LAMA-UHN accuracy of CV/MDL-chosen prompts vs. accuracy of the worst, average (randomly-selected), and best prompt (prior work). \textbf{Right}: The average accuracy gain from using CV/MDL-chosen prompts instead of randomly-chosen ones, relative to the gain from the best prompt. We plot mean/std. err. across 5 runs with different training sets. Across all model sizes, CV/MDL-chosen prompts obtain only small improvements over randomly-chosen ones and perform far worse than the best prompts.}
    \label{fig:plot_results_by_engine}
\end{figure*}

Fig.~\ref{fig:plot_results_by_engine} (left) shows the results; prompt selection obtains marginal improvements over random selection across model sizes ranging over 3 orders of magnitude. Prompts chosen by CV and MDL alike underperform the best prompt (chosen using held-out performance) by 5-7\% absolute on average. In fact, prompts chosen based on held-out performance often outperform larger models whose prompts are chosen in a true few-shot manner. CV and MDL do tend to choose better-than-average prompts, but only close the gap between the average and best prompts by 20-40\%, as shown in Fig.~\ref{fig:plot_results_by_engine} (right).

Fig.~\ref{fig:plot_results_distribution} (left) shows that CV/MDL struggle to choose the prompt with the highest test accuracy. Poor top-prompt selection is especially prevalent for larger models like GPT-3 175B that have spurred interest in prompt design (only 21\% accuracy for CV vs. 9\% for random chance). Altogether, our results show that effective prompt selection is difficult in the true few-shot setting, and that prior work overestimated the ability of LMs by using held-out examples for prompt selection.

\subsection{How reliably does prompt selection improve over the average prompt?}
\label{ssec:How reliably does prompt selection improve over the average prompt?}
\begin{figure*}[t]
	\centering
    \includegraphics[scale=.33]{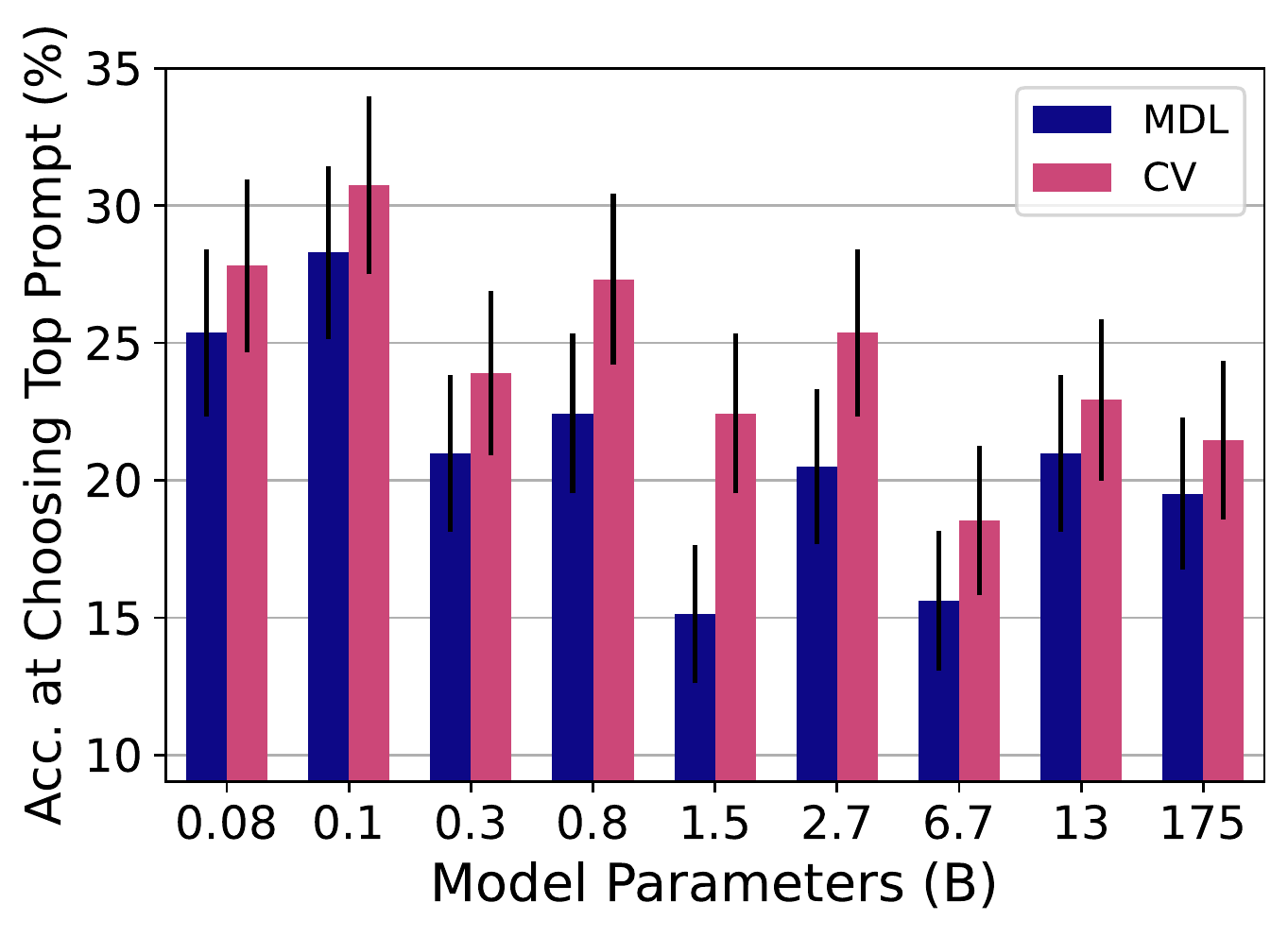}
    \includegraphics[scale=.33]{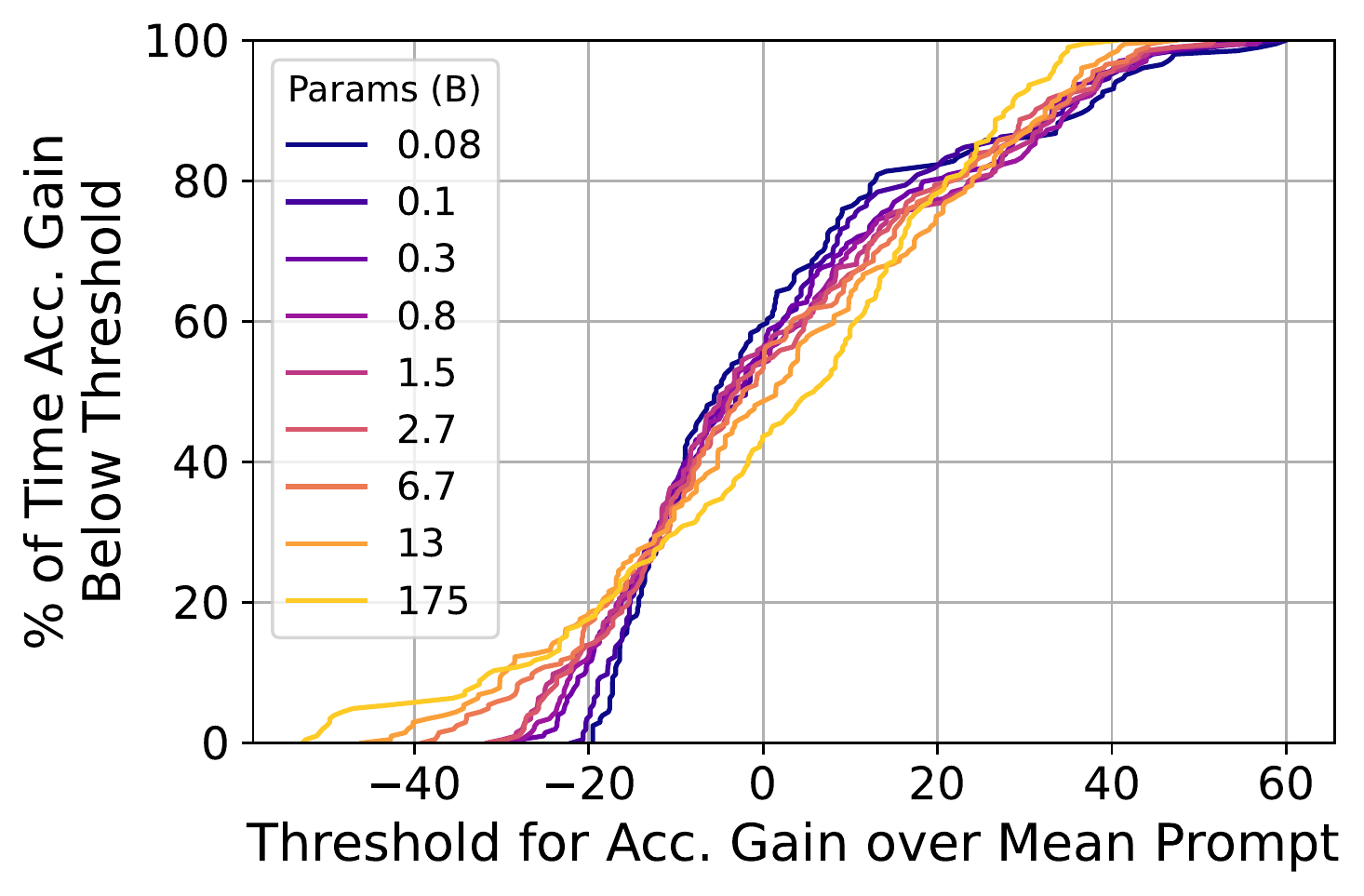}
    \includegraphics[scale=.33]{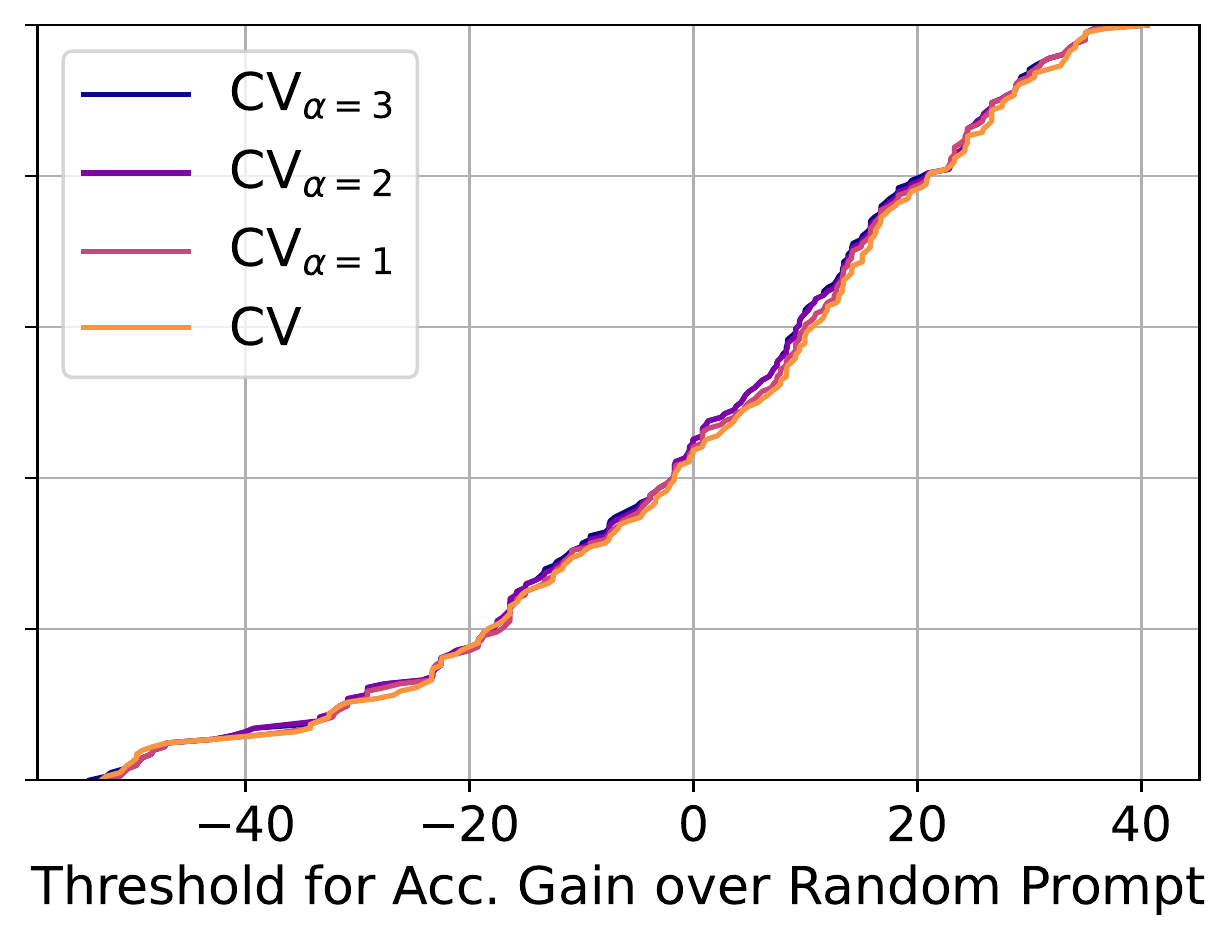}
    \caption{\textbf{Left}: CV/MDL have low accuracy at choosing the best prompt (mean/std. err. across 5 runs with different training sets). \textbf{Middle}: The chance of various accuracy gains on LAMA over the average prompt, when using prompts chosen by CV, and (\textbf{Right}) conservative estimates of CV that also minimize variance in CV;
    CV often chooses worse-than-average prompts, an issue that is not mitigated with conservative prompt selection.
    }
    \label{fig:plot_results_distribution}
\end{figure*}

If the expected improvement from prompt selection is small, can we at least obtain an improvement with high probability for any given task and training set?
Fig.~\ref{fig:plot_results_by_engine} (left) shows that the worst prompts perform far worse than average, so it would be useful if prompt selection helped to avoid the worst prompts.
We examine the probability with which prompt selection obtains various accuracy gains over the average (randomly-chosen) prompt and show results in Fig.~\ref{fig:plot_results_distribution} (middle) for CV (and similar results in Appendix \S\ref{sec:Additional Results with MDL} for MDL).

CV/MDL-chosen prompts show high variance in test accuracy relative to the average prompt.
For most model sizes (.1B-6.7B), the chance of improving over the average, randomly-chosen prompt is only $\sim$56\% for CV and $\sim$55\% for MDL.
The performance of prompt selection forms a long-tailed distribution; there is a $\sim$27\% chance that prompt selection causes an accuracy drop of $\sim$13\% for all model sizes and CV/MDL alike.
Furthermore, the tails grow heavier as model size increases.
For the largest model (GPT-3 175B), CV/MDL-chosen prompts sometimes do far worse than average, e.g., 40\% worse, 5\% of the time.
Our results suggest a troubling trend: as models grow bigger and generalize better, our ability to reliably choose good prompts degrades.
One possible explanation is that larger models have the capacity to draw more complex decision boundaries, requiring more examples to estimate the true expected loss on unseen examples; we may need to scale validation sets along with model size.
Overall, the limited average-case gains from prompt selection cannot be expected with any reasonable confidence in the true few-shot setting, a problem that will only become worse with larger models.

\subsection{Can we increase the likelihood of improved performance from prompt selection?}
\label{ssec:Can we increase the likelihood of prompt selection to improve performance?}

As we have shown, CV and MDL do not reliably choose better-than-average prompts. Here, we explore the extent to which we can reduce the variance in generalization by explicitly preferring prompts with low variance (\S\ref{ssec:Variance Matters}).
For the largest model (GPT-3 175B), we choose prompts based on a conservative estimate of generalization loss, $\text{CV}_{\alpha}$ (\S\ref{ssec:Variance Matters}).
We show the test accuracy for the prompt chosen with various levels of confidence $\alpha \in \{1, 2, 3\}$ and with CV ($\alpha=0$).

As shown in Fig.~\ref{fig:plot_results_distribution} (right), all $\alpha$ lead to a similar distribution of performance gain as CV. For example, CV outperforms the average prompt 50\% of the time vs. 51\% for $\alpha=2$. These results suggest that it is non-trivial to choose prompts that reliably perform better than random selection, even when explicitly minimizing variance in generalization, further highlighting the difficulty of reliably selecting good prompts in the true few-shot setting.

\subsection{Does prompt selection improve with more labeled examples?}
\label{ssec:Does prompt selection improve with more labeled examples?}
\begin{figure*}[t]
	\centering
        \includegraphics[scale=.40]{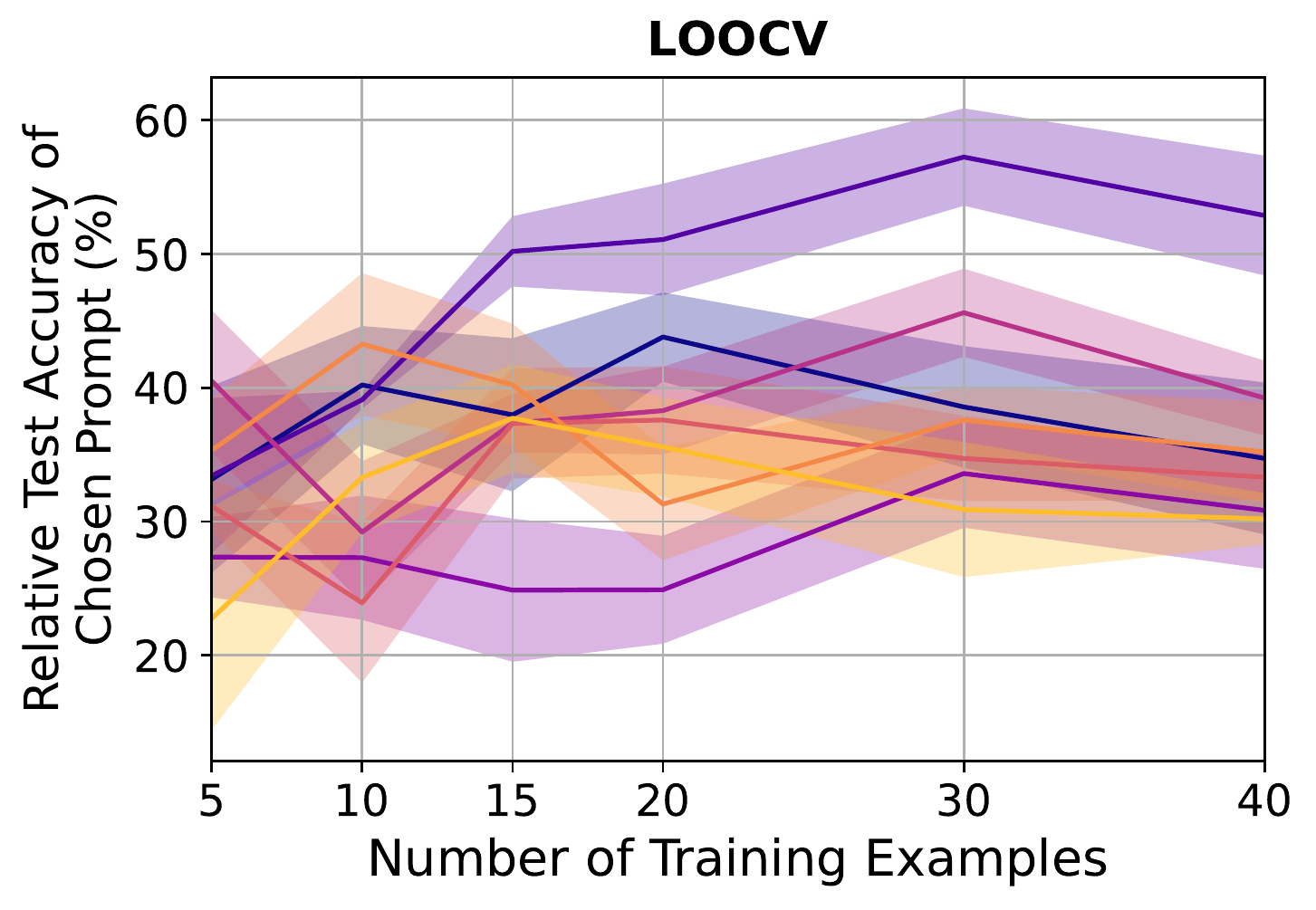}
        \includegraphics[scale=.40]{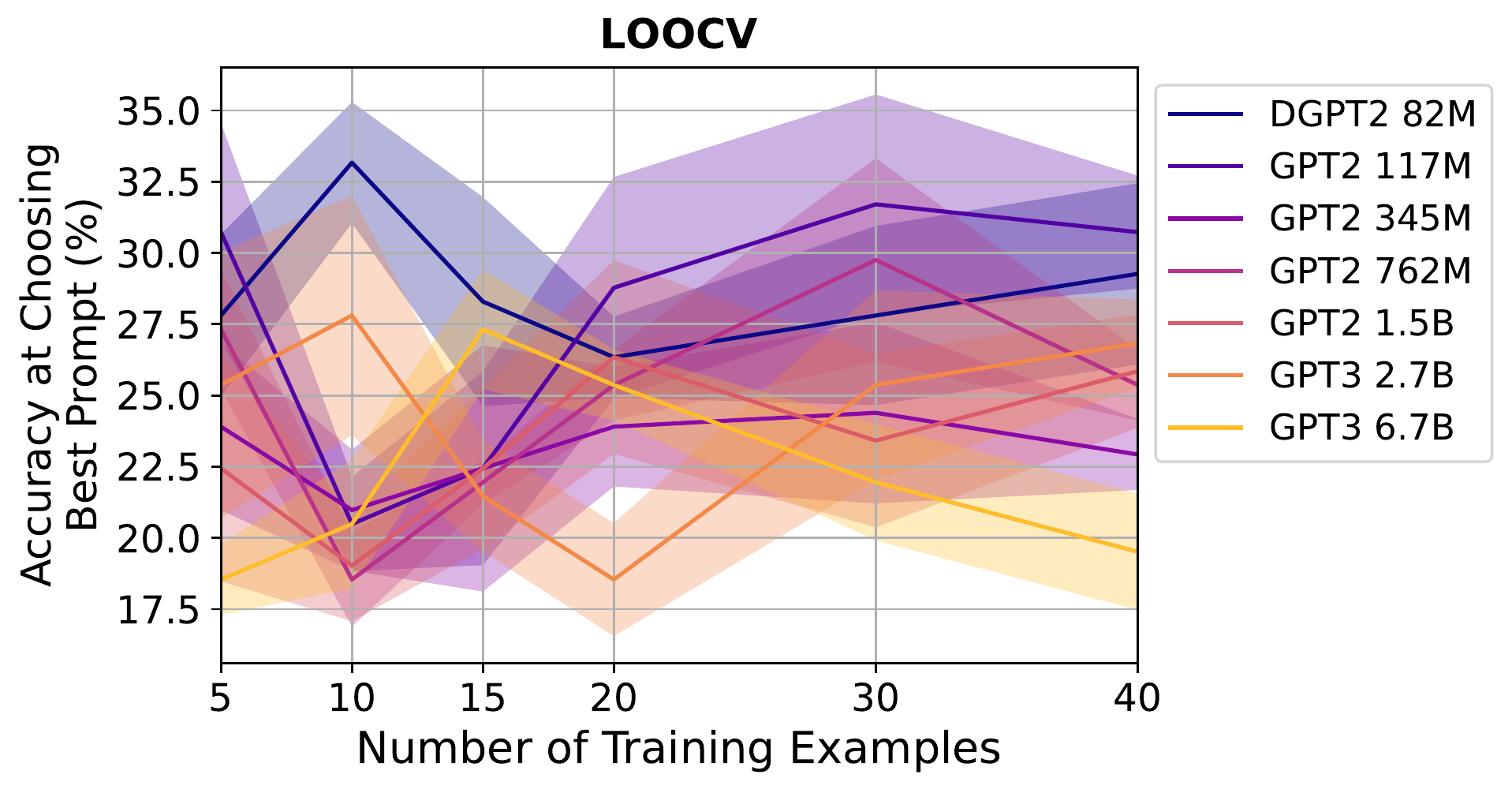}
    \caption{Increasing the number of examples up to 40 does not clearly improve CV in terms of (\textbf{Left}) accuracy gain over the average prompt (scaled to 0), relative to the best one (scaled to 100) or (\textbf{Right}) accuracy at choosing the best prompt. Mean/std. err. on LAMA over 5 runs (varying train sets).}
    \label{fig:plot_results_by_num_train-loo}
\end{figure*}

The poor performance of prompt selection methods may be due to using such a small number of labeled examples. As the number of labeled examples increases, we expect prompt selection methods to improve.
Thus, true few-shot prompt selection may be possible with a few dozen examples (though it is not always possible to use more examples, due to limits on input length for LMs like GPT).
We therefore examine the test accuracy of CV/MDL-chosen prompts as we use an increasing number of labeled examples $N \in \{5, 10, 15, 20, 30, 40\}$.
For $N \geq 10$, it is not feasible to marginalize over all possible training example permutations, so we randomly sample 120 permutations (to match $N=5$) such that each example occurs the same number of times in each position (i.e., to use each example as the held-out CV fold the same number of times).
We run the experiment for $\leq$6.7B parameter models, since it is prohibitively costly to run with larger models via the OpenAI API.

As shown in Fig.~\ref{fig:plot_results_by_num_train-loo}, there is no consistent trend in the performance of prompt selection, both in terms of task performance (left) and in terms of accuracy at choosing the highest accuracy prompt (right). Even in higher-data regimes (40 examples), CV/MDL struggle to choose effective prompts and do not consistently, across model sizes, perform better than choosing examples based on 5 examples. Our findings are surprising, because the true-few shot setting is where prompt design has been thought most promising, due to the scarcity of training data~\citep[][]{scao2021data}. However, the true few-shot setting is also one in which prompt selection is hardest, greatly undermining the potential value of prompts.

\subsection{Does prompt selection improve with more computation?}
\label{ssec:Does prompt selection improve with more computation?}
\begin{figure*}[t]
	\centering
    \includegraphics[scale=.33]{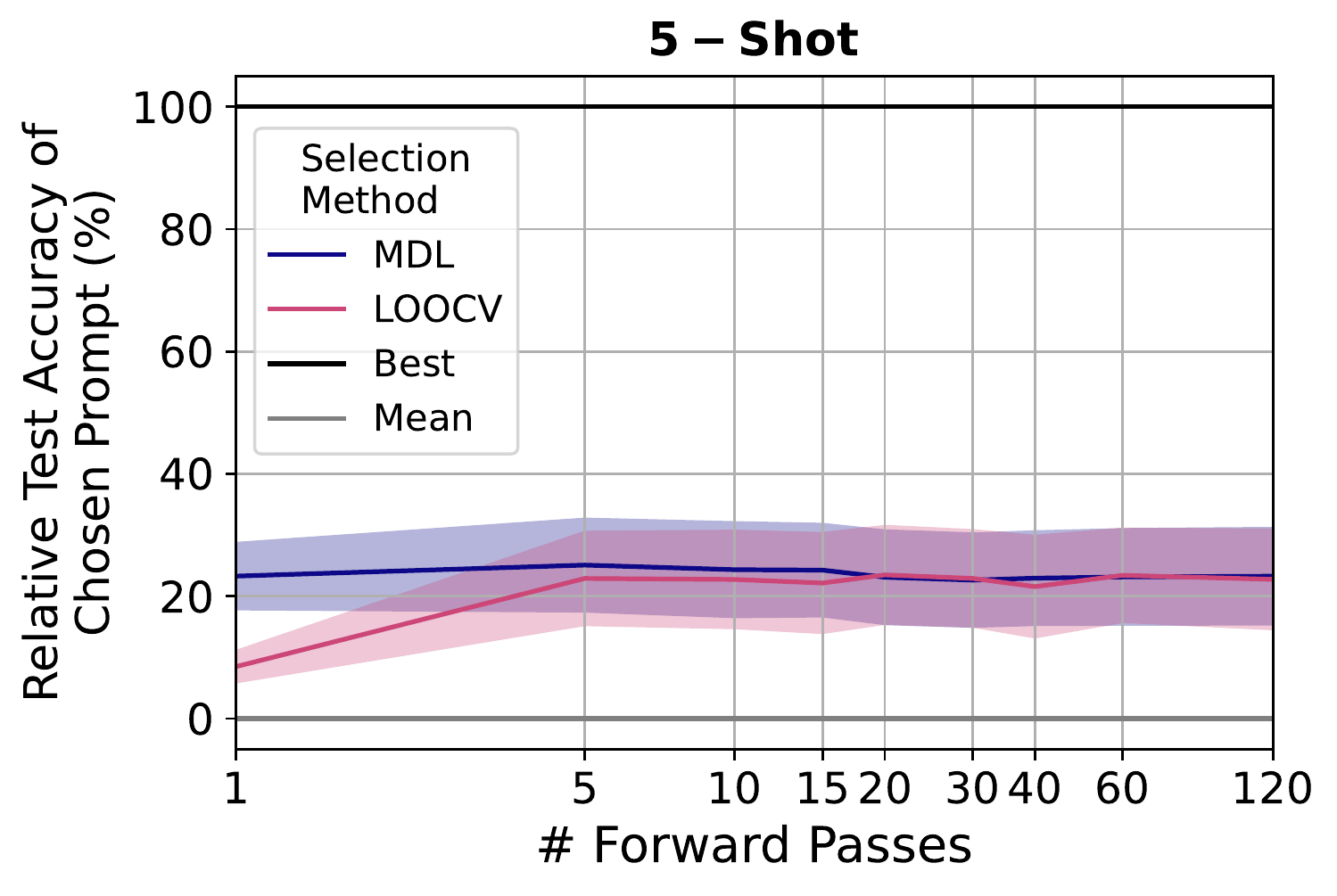}
    \includegraphics[scale=.33]{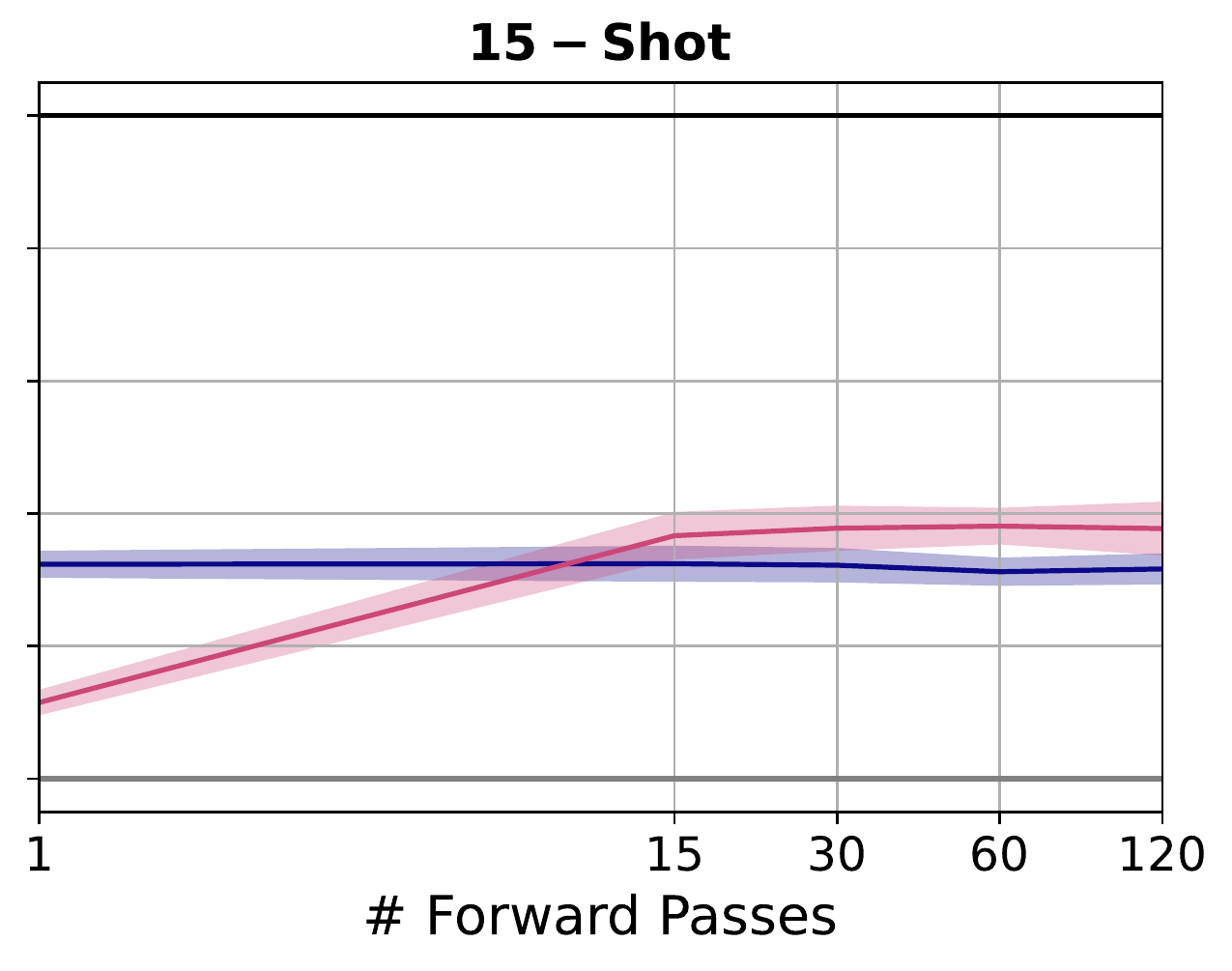}
    \includegraphics[scale=.33]{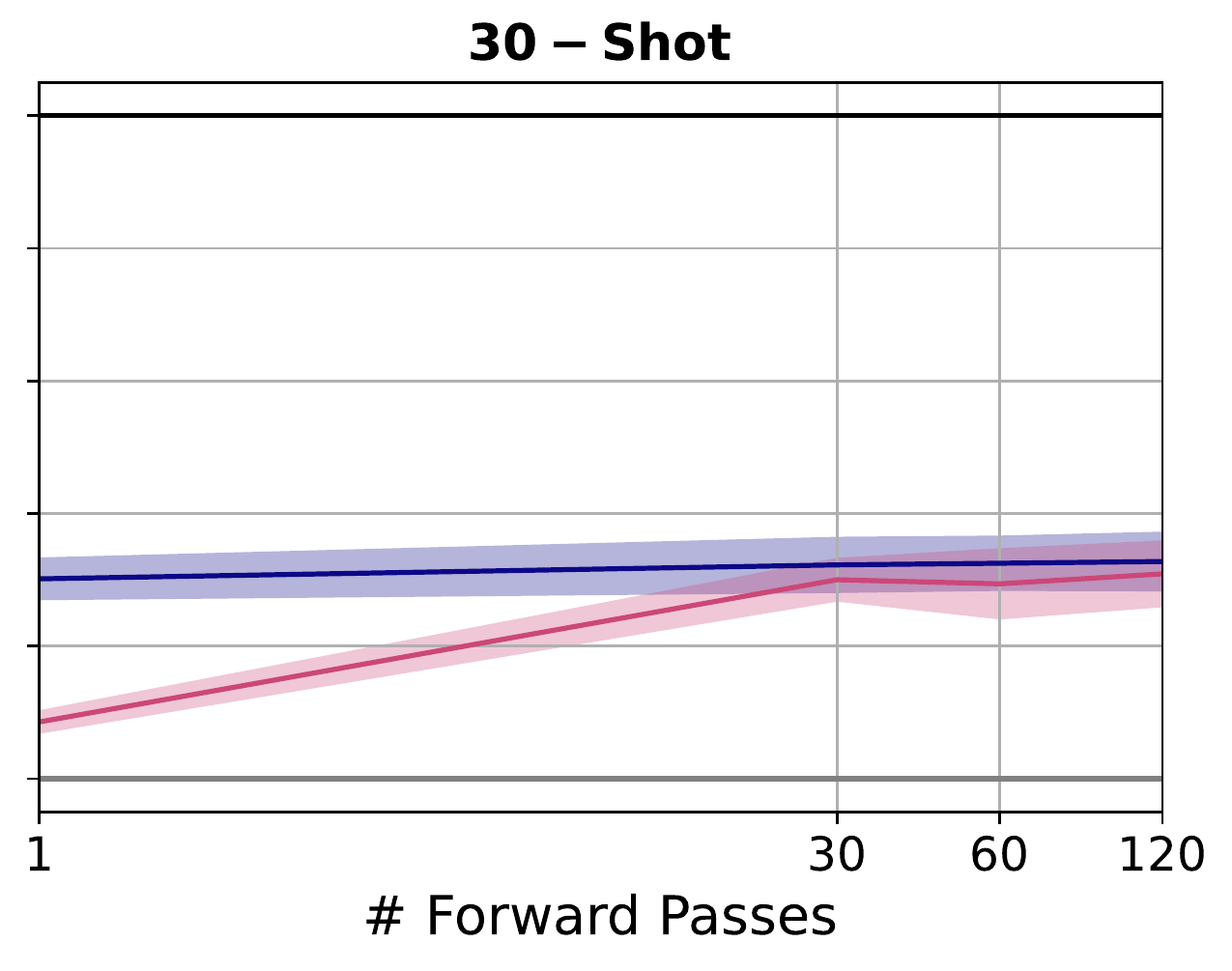}
    \caption{For $N \in \{5,10,30\}$ -shot learning, increasing the compute used to estimate CV/MDL does not notably improve the accuracy of chosen prompts beyond a certain point (1 forward pass for MDL, $N$ forward passes for CV). Mean/std. err. across 5 runs for GPT-3 6.7B.}
    \label{fig:plot_compute_efficiency}
\end{figure*}

In the preceding sections, we computed $\mathbb{E}_{R,F}[\text{CV}(\mathcal{A}, R, F)]$ using a fixed number of samples for $R$.
Can we improve prompt selection by using more samples, at the cost of increased computation?
To answer this question, we vary the number of samples of $R$ (and thus LM forward passes) used to compute the above expectation and choose prompts as described in \S\ref{ssec:Variance Matters}.
To estimate CV with a single forward pass, we sample a single fold $k$ (here, a single example) and evaluate accuracy on fold $k$ when conditioning the LM on all others folds.
Fig.~\ref{fig:plot_compute_efficiency} shows the results for $N \in \{5, 15, 30\}$ training examples using the largest model from \S\ref{ssec:Does prompt selection improve with more labeled examples?} (GPT-3 6.7B).

Computation is not the bottleneck in prompt selection, as test accuracy roughly plateaus after one forward pass for MDL and $N$ forward passes for CV.
This observation holds across $N$, as well as all models with $<$6.7B parameters (omitted for space).
Our results suggest that true few-shot prompt selection is fundamentally limited by the number of examples available.

\subsection{To what extent are chosen prompts specific to the model?}
\label{ssec:To what extent are chosen prompts specific to the model?}
\begin{figure*}[t]
	\centering
    \includegraphics[scale=.49]{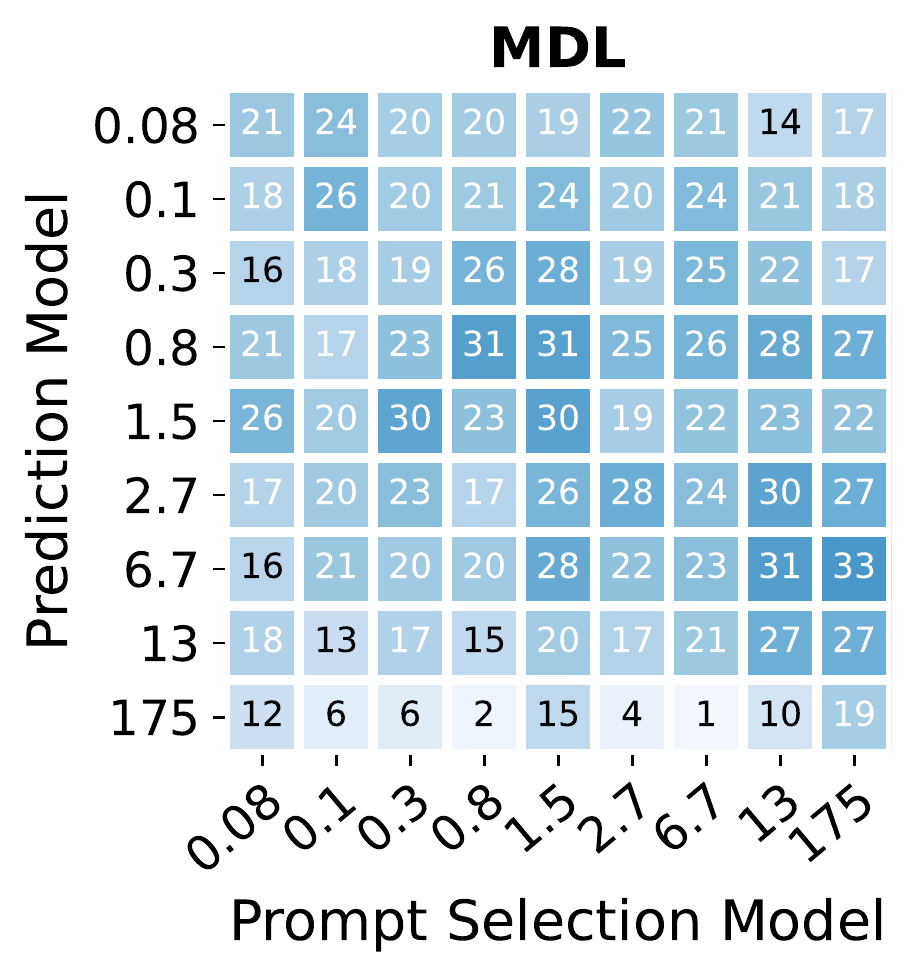}
    \includegraphics[scale=.49]{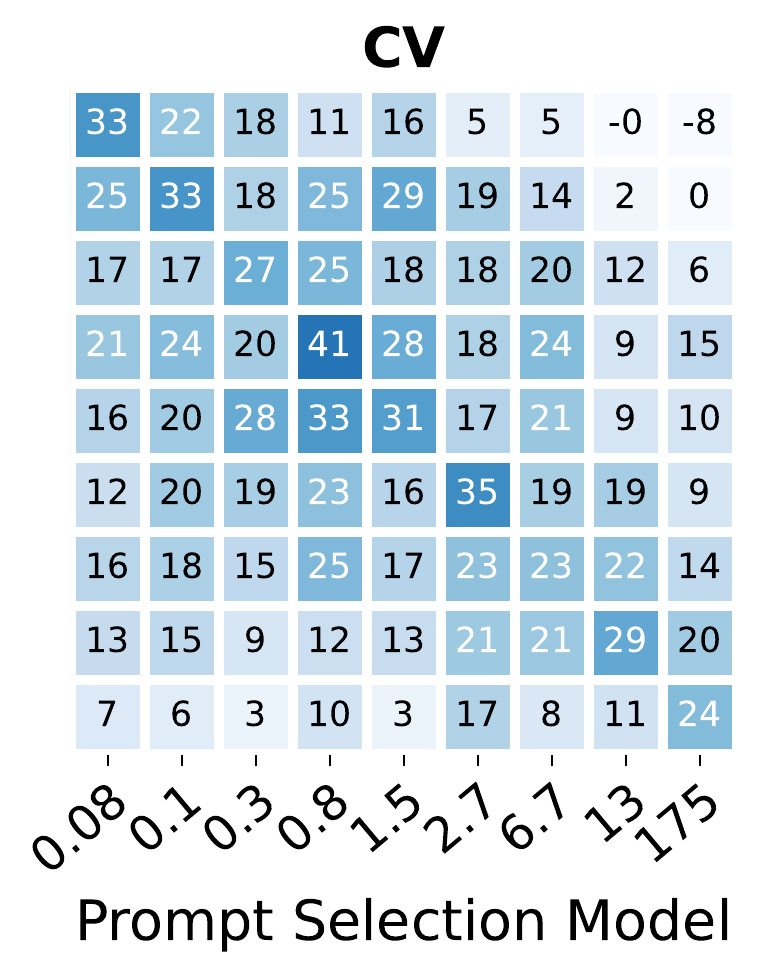}
    \includegraphics[scale=.49]{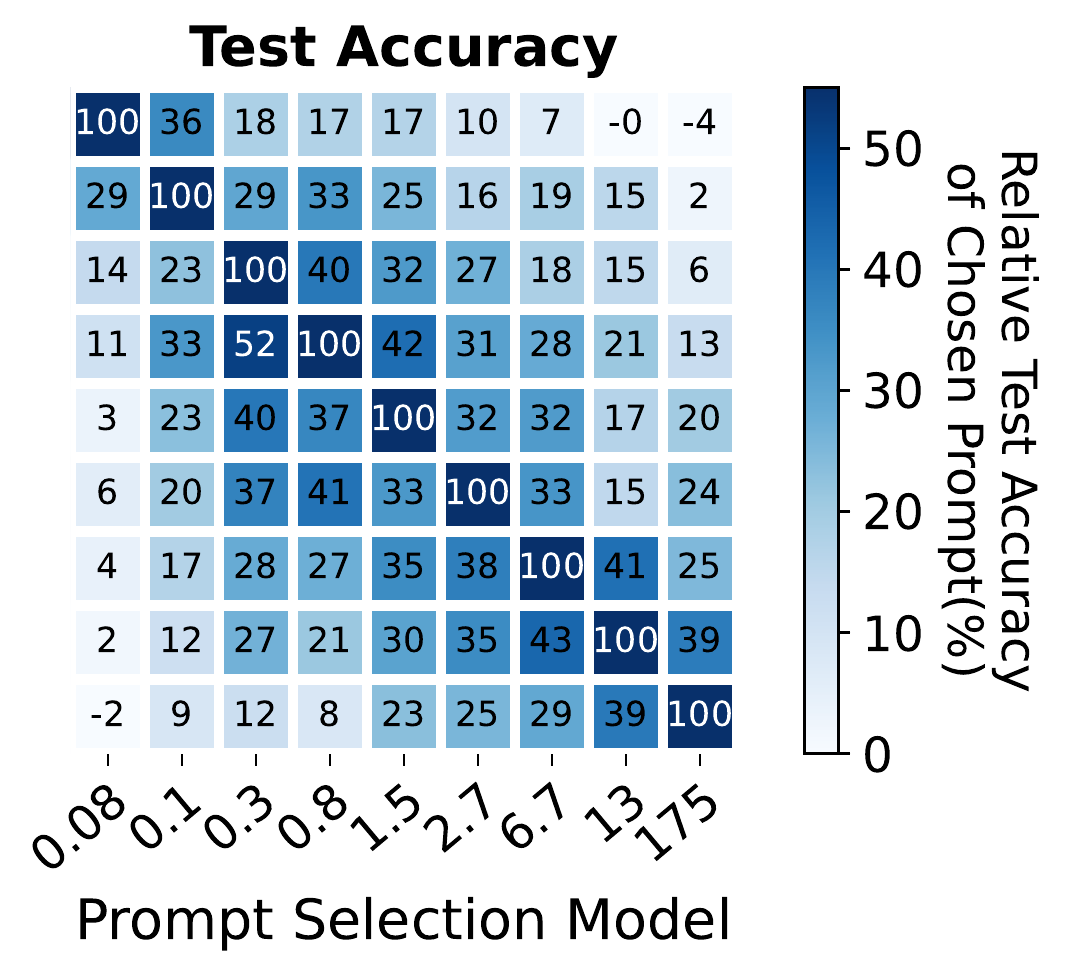}
    \caption{
    A model's accuracy with the prompt chosen for another model using MDL, CV, or test accuracy. We show LAMA accuracy relative to the average prompt (scaled to 0) and best prompt (scaled to 100) for a model size. CV/MDL show different patterns in prompt transfer than test acc.}
    \label{fig:plot_prompt_transfer}
\end{figure*}

We investigate the extent to which CV/MDL-chosen prompts differ from the best, test-chosen prompts in other ways, aside from accuracy.
To this end, we examine how well a model does when using a prompt chosen for another model, which we refer to as ``prompt transfer.''
Prompt transfer indicates how tailored the chosen prompt is to a given model.
For each model, we examine the average gain of the chosen prompt over the average prompt, relative to the maximum possible gain, i.e., scaling the test accuracy for each model so that the average prompt scores 0\% and the top prompt scores 100\%.

As shown in Fig.~\ref{fig:plot_prompt_transfer}, prompts chosen based on test accuracy generalize reasonably well across models of similar sizes, a pattern that degrades as we examine CV and especially MDL.
For CV, prompts chosen using one model size do transfer better to similar model sizes, but CV-chosen prompts do not transfer as effectively as test-chosen ones.
For MDL, the chosen prompts are not particularly tailored to the given model, performing similarly across many model sizes.
Overall, even the pattern of prompt transfer differs between test accuracy and CV/MDL.

\begin{figure*}[t]
	\centering
    \includegraphics[scale=.34]{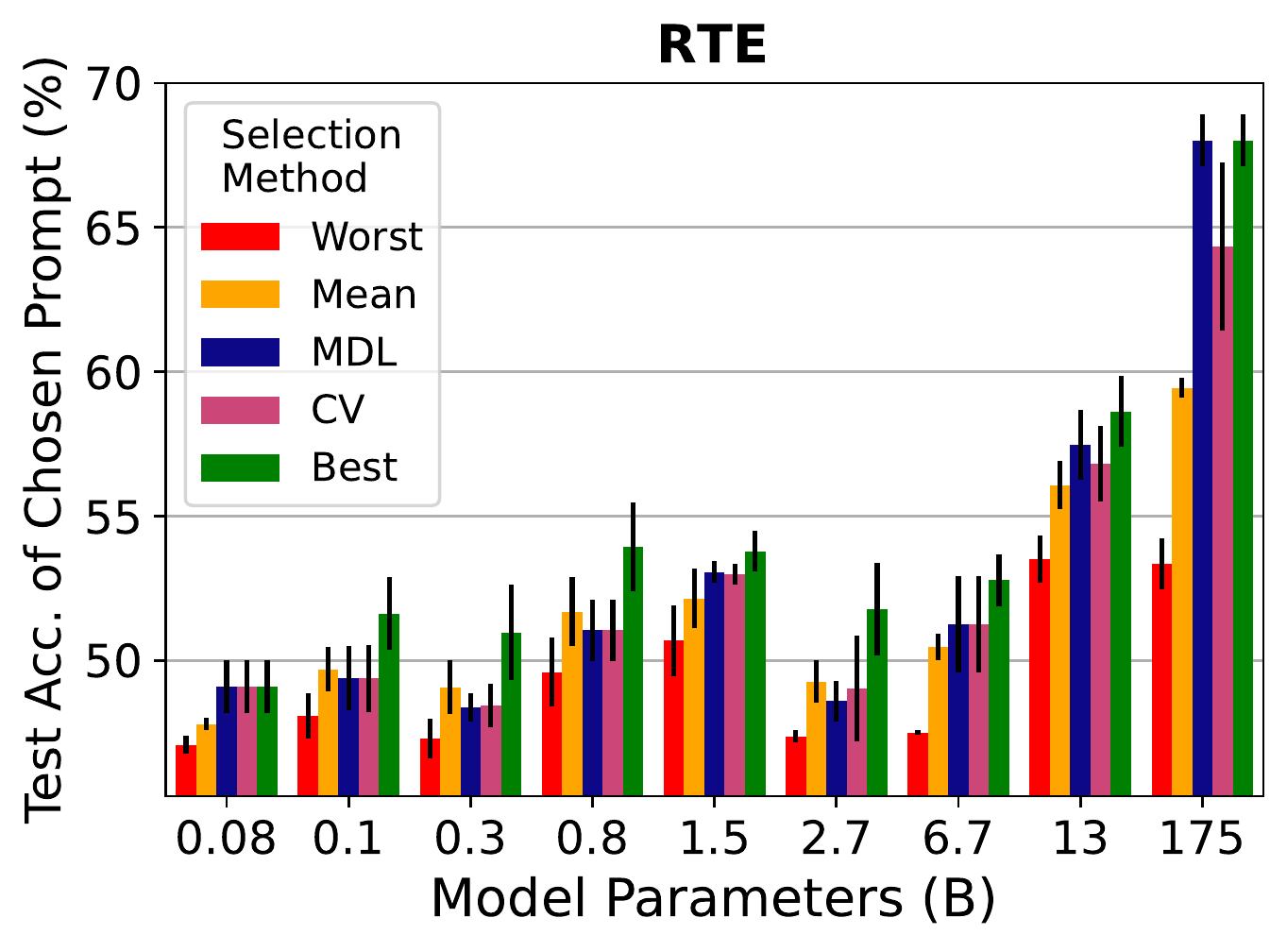}
    \includegraphics[scale=.34]{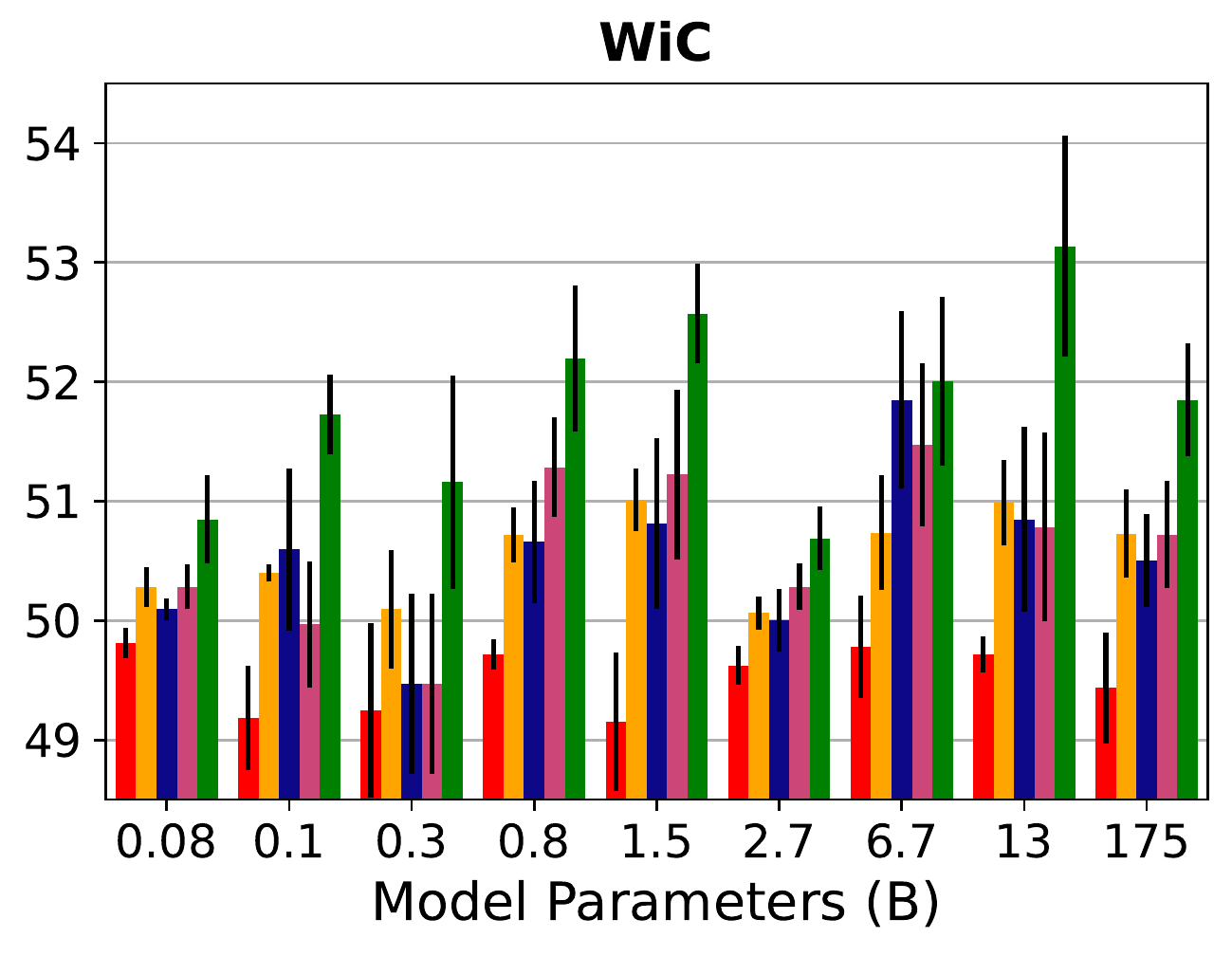}
    \includegraphics[scale=.34]{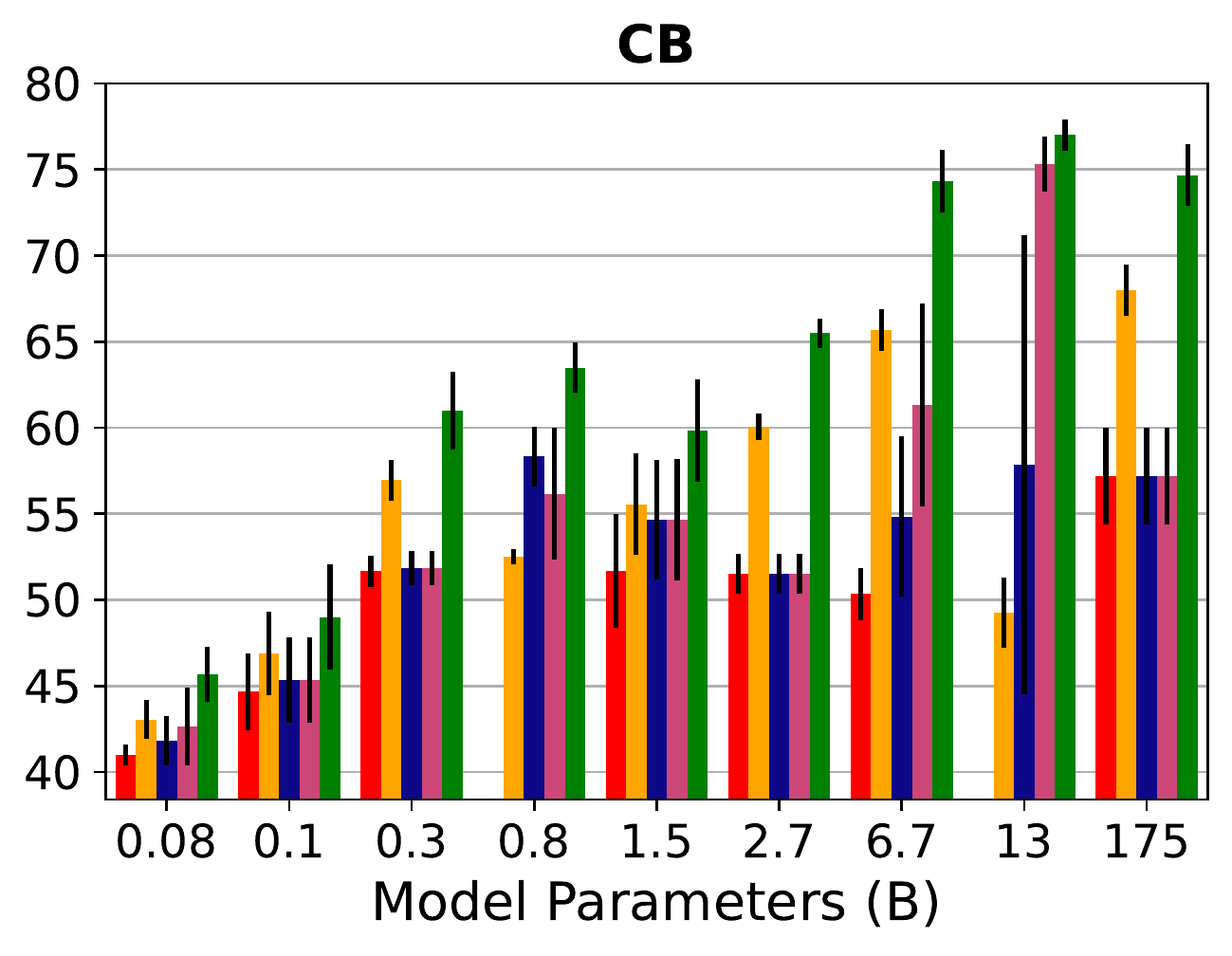}
    \caption{Accuracy of CV/MDL-chosen prompts vs. accuracy of the worst, average (randomly-selected), and best prompt (prior work), on three classification tasks (mean/std. err. over 5 runs). CV/MDL-chosen prompts generally perform several points worse than the best prompt and do not consistently improve over the average prompt across tasks and model sizes.}
    \label{fig:plot_results_by_engine-superglue}
\end{figure*}

\begin{figure*}[t]
	\centering
    \includegraphics[scale=.35]{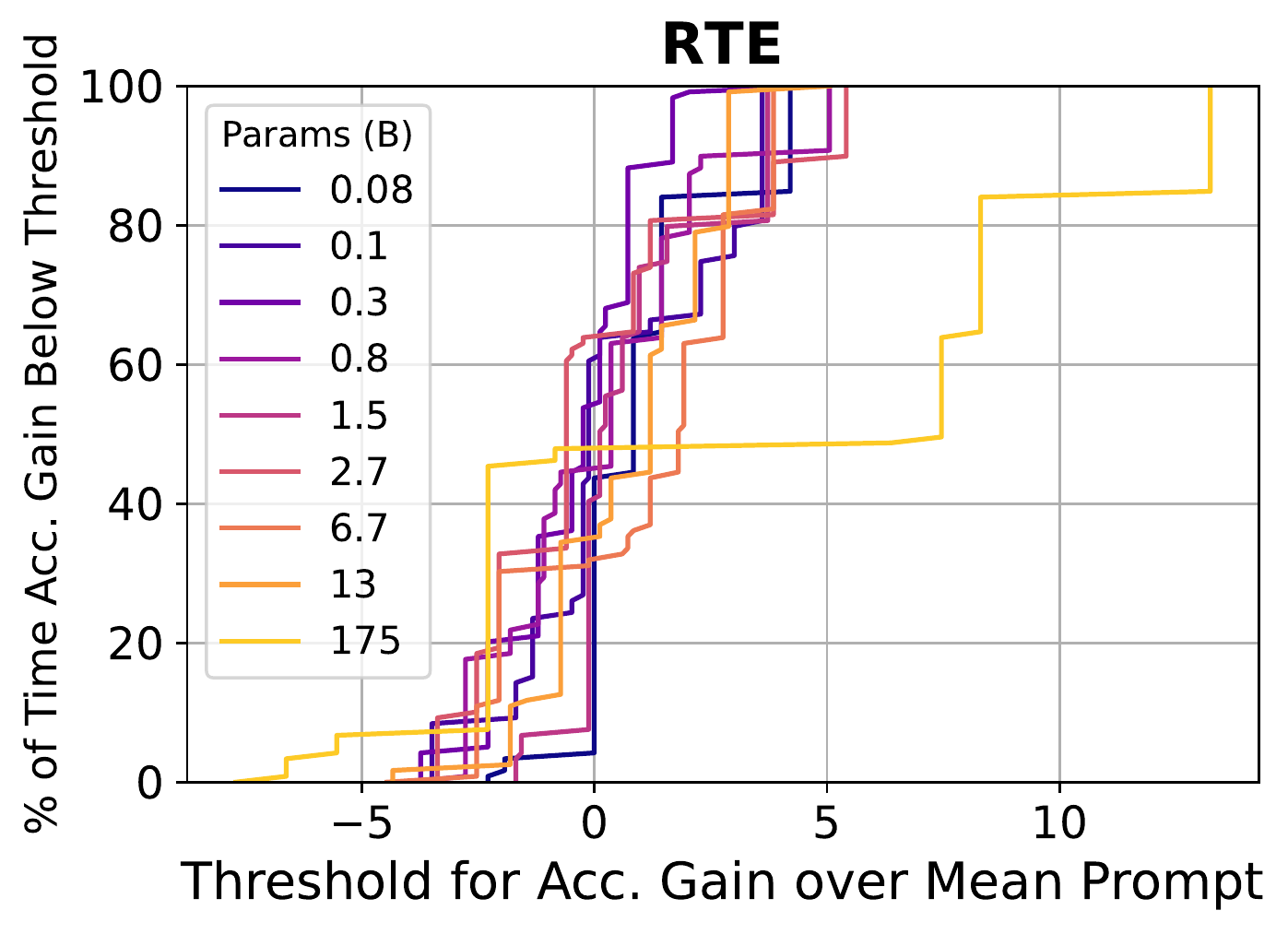}
    \includegraphics[scale=.35]{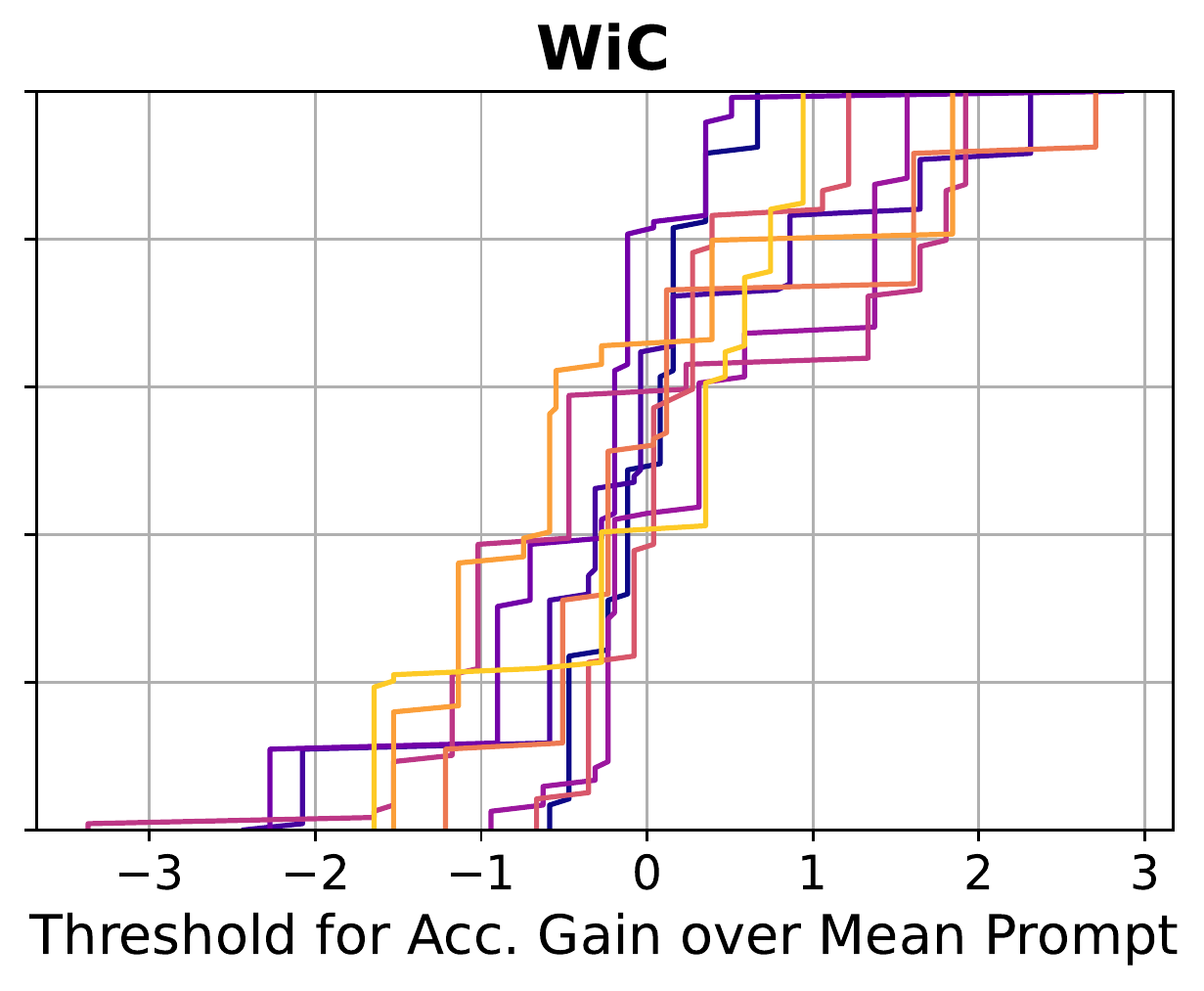}
    \includegraphics[scale=.35]{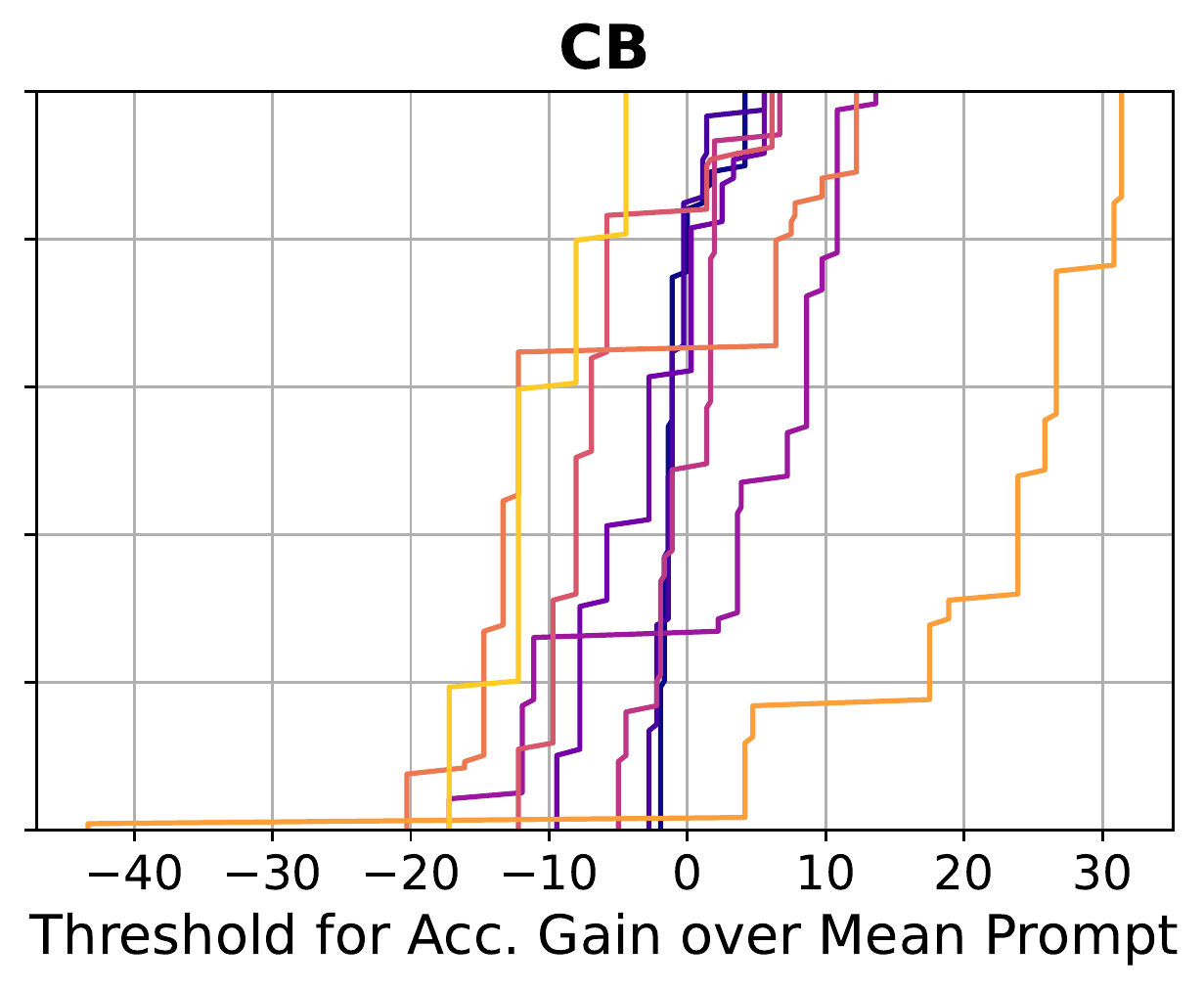}
    \caption{The chance of various accuracy gains over the average prompt from CV on RTE, WiC, and CB.
    CV often chooses prompts that are below average (RTE, WiC) or far below average (CB).}
    \label{fig:plot_results_distribution-superglue}
\end{figure*}

\subsection{Is prompt selection challenging on other tasks?}
\label{ssec:Is prompt selection challenging on other tasks?}
We now examine the extent to which our results on LAMA tasks hold on other kinds of NLP tasks.
We examine three classification tasks for which prior work has designed various prompts: Recognizing Textual Entailment (RTE), CommitmentBank (CB), and Word-in-Context (WiC).
RTE and CB involve detecting whether one sentence entails or contradicts another, and WiC involves determining if a  polysemous word is used with the same sense in two sentences (e.g., ``Room and \underline{board}'' and ``He nailed \underline{boards} across the windows.''); See Appendix\S\ref{ssec:SuperGLUE} for further task details.
We evaluate the accuracy of GPT models when using prompts chosen by CV, MDL, and test accuracy, as we did for LAMA.
For each task, we evaluate held-out accuracy using the full validation set when using 5 training examples randomly sampled from the task train set, while ensuring that we include at least one example per class.
We evaluate the mean/std. error over 5 train sets.
As our set of prompts, we use the manually-written prompts from~\cite{brown2020language} and~\cite{schick2020small} -- 3 prompts for RTE/CB and 4 prompts for WiC.
Schick and Schütze \cite{schick2020small} designed prompts for bidirectional LMs, so when necessary, we modify their prompts to be suitable for left-to-right LMs (see Appendix \S\ref{ssec:SuperGLUE} for prompts).
Fig.~\ref{fig:plot_results_by_engine-superglue} shows the accuracy of the chosen prompts on each task.

We observe as similar trend as before, that across tasks and model sizes, the CV/MDL-chosen prompt almost always obtains lower average accuracy than choosing based on test accuracy.
The trend holds even when choosing between fewer prompts (here, 3-4).
CV/MDL-chosen prompts vary greatly in test accuracy across tasks and model sizes, often choosing worse-than-average prompts (e.g., on CB).

We examine the variance in chosen prompt accuracy in more detail, by showing the chance that selection obtains various accuracy gains over the average prompt.
Here, we choose prompts with CV using $N$ forward passes (one evaluation per fold), as it represents a good tradeoff between compute and accuracy that is likely to be used in practice.
As shown in Fig.~\ref{fig:plot_results_distribution-superglue}, accuracy gains are again highly dispersed, often negative, and not consistently achieved.
For CB, there is a 20\% change of a 15\% accuracy drop for GPT-3 175B.
Model sizes vary greatly in how often the CV-chosen prompt leads to improvement, e.g., from 38-82\% for WiC and 1-83\% for CB.
Overall, our earlier findings carry over to other kinds of tasks, indicating that prompt selection is challenging in general.

\section{True Few-Shot Hyperparameter Selection}
\label{sec:True Few-Shot Hyperparameter Selection}

Having shown that true few-shot prompt selection is challenging, we now study the effectiveness of model selection methods in the context of hyperparameter selection more generally.
As our model, we examine ADAPET~\citep{tam2021improving}, as it is open-source\footnote{\url{https://github.com/rrmenon10/ADAPET}} and currently the top-performing few-shot model according to SuperGLUE~\cite{wang2019superglue}, a standard benchmark in NLP.
ADAPET finetunes the pretrained $\text{ALBERT}_{\text{xxlarge-v2}}$ LM~\citep{lan2020albert} to (1) classify each label as correct or incorrect given the input and (2) to predict randomly masked out input tokens given the label and unmasked input tokens, similar to Masked LM~\citep{devlin-etal-2019-bert}.
ADAPET was developed in the context of tuned few-shot learning, as ADAPET's hyperparameters were chosen based on generalization to validation examples.
We investigate how ADAPET does in the true few-shot setting.

We evaluate the impact of using validation examples to choose two hyperparameters: the early stopping checkpoint and fraction of words masked for the masked LM objective.
ADAPET performs $T=1000$ gradient updates on batches of 16 examples and chooses the checkpoint at $T \in \{250, 500, 750, 1000\}$ with the highest validation accuracy.
ADAPET also chooses the best masking fraction $M \in \{0.075, 0.10, 0.105, 0.15\}$.
Following ADAPET, we evaluate on SuperGLUE, a suite of 8 NLP tasks.
SuperGLUE consists of four question-answering tasks (BoolQ, COPA, MultiRC, ReCoRD), a coreference resolution task (WSC), as well as WiC, RTE, and CB discussed in \S\ref{ssec:Is prompt selection challenging on other tasks?} (see Appendix \S\ref{ssec:SuperGLUE} for task details).
We use CV/MDL to choose $T$ and $M$ (out of 16 total combinations) and then train a model on the full dataset with the chosen $T$ and $M$.
We use FewGLUE~\citep{schick2020small}, the 32-example subset of SuperGLUE used in prior work on few-shot learning.
We also use 3 other 32-example subsets that we randomly sample from SuperGLUE, to estimate variance in performance across training sets.
ADAPET uses a prompt during fine-tuning, choosing the prompt based on validation examples.
To avoid using validation-tuned prompts, we use the first prompt for every task as the authors do for ablation studies.
Since training ADAPET is expensive, we evaluate CV/MDL with $K=8$ folds.\footnote{See Appendix \S\ref{ssec:Computing MDL with ADAPET} for details on how we evaluate MDL on different SuperGLUE tasks.}
We show results in Table~\ref{tab:adapet}.

\paragraph{Results}

\begin{table*}[t]
\centering
\resizebox{1\textwidth}{!}{%
\begin{tabular}{lcccccccc|c}
\toprule
{} & \textbf{BoolQ} &                 \textbf{CB} & \textbf{COPA} &  \textbf{RTE} &  \textbf{WiC} &  \textbf{WSC} &           \textbf{MultiRC} &            \textbf{ReCoRD} &  \textbf{Avg} \\
{} &            Acc &                      Acc/F1 &           Acc &           Acc &           Acc &           Acc &                      EM/F1 &                      EM/F1 &               \\
\midrule
\textbf{Worst } &   75.0$_{4.8}$ &   79.5$_{2.3}$/67.3$_{7.8}$ &  76.8$_{2.2}$ &  63.2$_{4.0}$ &  49.0$_{1.3}$ &  77.2$_{1.8}$ &  38.5$_{7.4}$/80.0$_{2.9}$ &  76.2$_{1.8}$/86.5$_{1.2}$ &  69.4$_{1.5}$ \\
\midrule
\textbf{Mean  } &   79.0$_{1.5}$ &  85.9$_{2.3}$/74.5$_{11.0}$ &  81.1$_{2.9}$ &  70.8$_{2.5}$ &  51.5$_{1.8}$ &  82.5$_{2.7}$ &  44.2$_{6.6}$/82.3$_{2.7}$ &  78.3$_{1.3}$/87.8$_{0.8}$ &  73.9$_{1.2}$ \\
\textbf{MDL   } &   76.5$_{5.8}$ &  85.7$_{5.6}$/74.8$_{13.4}$ &  82.0$_{2.9}$ &  70.4$_{8.5}$ &  52.2$_{3.0}$ &  82.0$_{3.1}$ &  39.7$_{8.1}$/80.6$_{3.2}$ &  78.9$_{0.7}$/88.2$_{0.4}$ &  73.4$_{2.8}$ \\
\textbf{CV    } &   78.9$_{2.4}$ &  83.9$_{5.3}$/69.2$_{10.3}$ &  80.5$_{3.3}$ &  68.7$_{7.0}$ &  51.1$_{1.6}$ &  83.1$_{2.6}$ &  41.9$_{7.2}$/81.4$_{3.1}$ &  78.7$_{1.6}$/88.1$_{1.0}$ &  73.0$_{2.1}$ \\
\midrule
\textbf{Best  } &   80.9$_{1.0}$ &  89.8$_{3.1}$/79.8$_{13.4}$ &  84.8$_{4.5}$ &  76.7$_{1.8}$ &  54.1$_{2.3}$ &  86.6$_{1.8}$ &  46.8$_{6.9}$/83.4$_{2.9}$ &  80.4$_{1.1}$/89.2$_{0.7}$ &  77.2$_{0.9}$ \\
\midrule
\textbf{ADAPET}~\citep{tam2021improving} & 80.3 & 89.3 / 86.8 & 89.0 & 76.5 & 54.4 & 81.7 & 39.2 / 80.1 & 85.4 / 92.1  &  77.3 \\
\textbf{iPET}~\citep{schick2020small}   & 80.6 & 92.9 / 92.4 & 95.0 & 74.0 & 52.2& 80.1 & 33.0 / 74.0 &  86.0 / 86.5 & 76.8 \\ 
\textbf{PET}~\citep{schick2020small}    & 79.4& 85.1 / 59.4& 95.0 & 69.8 & 52.4 & 80.1 & 37.9 / 77.3 & 86.0 / 86.5  & 74.1 \\
\textbf{GPT-3}~\citep{brown2020language}  & 77.5 & 82.1 / 57.2 & 92.0 & 72.9 & 55.3 & 75.0 & 32.5 / 74.8 & 89.0 / 90.1 & 73.2 \\
\end{tabular}
}
\caption{\label{tab:adapet}ADAPET results on SuperGLUE validation when choosing early stopping checkpoint and masked LM rate using CV/MDL vs. the worst/mean/best hyperparameters chosen with validation (mean$_\text{std. dev.}$ over four 32-shot train sets). On all tasks, CV/MDL-chosen hyperparameters perform similar to or worse than average, and several points below the best hyperparameters.
}
\end{table*}

Across all SuperGLUE tasks, CV/MDL hyperparameter selection performs similar to or worse than average (randomly-chosen) hyperparameters and several points worse than the best hyperparameters.
In the true few-shot setting, the average SuperGLUE performance of ADAPET drops below that of earlier methods (PET and iPET), highlighting how the use of validation examples can give the false appearance of progress in few-shot learning.
On MultiRC, CV/MDL choose hyperparameters that give similar performance to the worst hyperparameters, another indication that model selection methods do not consistently prevent worst-case behavior in the true few-shot setting.
Preliminary analysis in Appendix \S\ref{sec:Additional Results with MDL} suggests that choosing better-than-average hyperparameters requires several thousand examples.
Overall, our results indicate that it is not just prompt selection but model selection in general that is challenging in very low-data regimes.

\section{Conclusion and Future Work}
\label{sec:Conclusion and Future Work}

Our work shows that it is challenging to make even the most basic decisions about few-shot learning algorithms using only a few labeled examples.
Instead, it may be more promising to make additional assumptions.
The meta-learning setting assumes access to data from many other tasks in order to perform learning and model selection~\citep{ravi2017optimization,li2017learning,triantafillou2020metadataset,ye2021crossfit}.
Transfer learning and multitask learning assume access to data that is directly related to the task with limited data~\citep{caruana1995learning,caruana1997multitask,phang2018stilts,liu-etal-2019-multi}.
Data augmentation techniques assume there is a viable way to create more data from limited data~\citep{kocijan-etal-2019-surprisingly,xie2020unsupervised,chen-etal-2020-mixtext,yang-etal-2020-generative}.
Other approaches assume unlabeled data and develop unsupervised model selection techniques~\citep{artetxe2018unsupervised-neural,lample2018unsupervised,perez-etal-2020-unsupervised}.
When labeled data is cheap, the simplest approach is to assume more examples for validation---in which case we might be better off training on the additional examples.
Unless we make such assumptions explicit, we cannot make meaningful comparisons between few-shot learning algorithms.
We find the above avenues to be more promising future directions than true few-shot learning given the challenge of model selection.

Inspired by prior work~\citep{oliver2018realistic,dodge-etal-2019-show}, we offer recommendations for future work in true few-shot learning:
\begin{itemize}
\item Report all hyperparameters (prompts) considered and the hyperparameter selection criteria.
\item Include validation examples in the number of examples used by a few-shot learning algorithm. Validation examples include all examples used to decide on any aspect of learning: hyperparameters, prompts, training objectives, decoding strategies, model architecture, etc.
\item Once you have decided on the learning algorithm, submit your model for test evaluation directly, without first evaluating on validation. Report the total number of test evaluations conducted (ideally, just one). Use the validation set only after test evaluation for any ablations you report, to avoid making decisions about your algorithm with the validation set.
\item Do not rely on hyperparameters from prior work that were tuned using validation examples for the same benchmark (e.g., SuperGLUE), to avoid indirectly benefiting from validation examples. Instead, re-tune such hyperparameters using only the given few-shot examples.
\end{itemize}
The above protocols are strict but mimic how a true few-shot learning algorithm would be used in a real, low-data setting.
To ensure researchers comply with such strict protocols, future benchmarks may need to keep large test sets private while releasing only a few labeled examples.

Given our negative results on true few-shot learning, a major question remains: is it possible to select models in a true zero-shot setting?
Prior work uses LMs for zero-shot learning by choosing an arbitrary prompt~\citep{petroni-etal-2019-language,ettinger-2020-bert} which requires no data but is suboptimal~\citep{jiang-etal-2020-know}.
Other efforts try multiple prompts and choose between them via trial and error alongside manual evaluation~\citep{radford2019language}, effectively leveraging human supervision.
CLIP~\citep{radford2021learning} achieves high zero-shot accuracy on ImageNet after extensively tuning prompts and label names using ImageNet's training set (1.28M examples), as we noticed from the open-source code.\footnote{\url{https://github.com/openai/CLIP/blob/main/notebooks/Prompt_Engineering_for_ImageNet.ipynb}}
The authors report a 5\% accuracy gain from tuning prompts alone, but the training examples used for tuning are not available in true zero-shot learning.
Without any labeled data, the problem of model selection is even more challenging than in the true few-shot case.
Overall, our work provides guidance for future work in few-shot learning by clarifying the assumptions made by the true few-shot setting and empirically demonstrates that model selection is a major roadblock to true few-shot learning.

\section{Limitations and Broader Impact}
\label{sec:Limitations and Broader Impacts}

We facilitate fair comparisons between few-shot methods in future work by disambiguating between three few-shot settings: multi-distribution, tuned, and true few-shot learning.
We highlight that one setting, tuned few-shot learning, gives up the practical advantage of few-shot learning by using many labeled examples.
Furthermore, we show that several tuned few-shot learning algorithms work significantly worse in the true few-shot setting, without tuning on many examples.
Our study is not exhaustive, however, and it is possible that effective true few-shot model selection is possible using other criteria (\S\ref{ssec:Other Model Selection Criteria}) or even heuristics not explored here.
In this event, our work will have discouraged work on a few-shot learning setting with applications to low-data settings, e.g., that involve low-resource languages or expert annotation.
Overall, however, we believe our work will redirect future work to few-shot settings with more practical applications.

We show that it is hard to detect when a small input change hurts an LM's generalization, even when the change appears reasonable to human readers.
We argue that practitioners will benefit from knowing such limitations, but they may also be discouraged from deploying LMs in many useful contexts, such as question-answering, hate speech detection, automatic translation, and commercial dialogue systems.
Our findings may also encourage adversaries to target LM-based applications and highlight which models are most susceptible to attack (e.g., larger models).
By shedding light on the shortcomings of (few-shot) LMs, we hope to spur future work to address these shortcomings.

\section*{Acknowledgments}
We are grateful to OpenAI for providing access and credits to GPT-3 via the API Academic Access Program, and we thank Miles Brundage, David Schnurr, Felipe Such, Ryan Lowe, and Ilya Sutskever for help with the API.
We thank GPT-3 authors Benjamin Mann and Gretchen Krueger for helpful feedback on our paper.
We thank Rakesh Menon for assistance with the ADAPET codebase, Shenglong Wang for cluster support, Zhengbao Jiang for LPAQA prompts, and Tal Linzen, Patrick Lewis, Eric Wallace, Adam Fisch, Stephen Roller, Aravind Rajeswaran, Gretchen Krueger, Amanda Ngo, Udit Arora, Sébastian Jean, Jason Phang, and the NYU NLP group for feedback on our draft.
KC is partly supported by Samsung Advanced Institute of Technology (Next Generation Deep Learning: from pattern recognition to AI) and Samsung Research (Improving Deep Learning using Latent Structure). KC also thanks Naver, eBay, NVIDIA, and NSF Award 1922658 for support. EP is grateful to NSF and Open Philanthropy for fellowship support.

\bibliographystyle{unsrt}
\bibliography{refs}

\begin{thebibliography}{10}

\bibitem{radford2019language}
Alec Radford, Jeff Wu, Rewon Child, David Luan, Dario Amodei, and Ilya
  Sutskever.
\newblock Language models are unsupervised multitask learners, 2019.

\bibitem{brown2020language}
Tom~B. Brown, Benjamin Mann, Nick Ryder, Melanie Subbiah, Jared Kaplan,
  Prafulla Dhariwal, Arvind Neelakantan, Pranav Shyam, Girish Sastry, Amanda
  Askell, Sandhini Agarwal, Ariel Herbert-Voss, Gretchen Krueger, Tom Henighan,
  Rewon Child, Aditya Ramesh, Daniel~M. Ziegler, Jeffrey Wu, Clemens Winter,
  Christopher Hesse, Mark Chen, Eric Sigler, Mateusz Litwin, Scott Gray,
  Benjamin Chess, Jack Clark, Christopher Berner, Sam McCandlish, Alec Radford,
  Ilya Sutskever, and Dario Amodei.
\newblock Language models are few-shot learners, 2020.

\bibitem{schick2020exploiting}
Timo Schick and Hinrich Schütze.
\newblock Exploiting cloze questions for few-shot text classification and
  natural language inference.
\newblock {\em Computing Research Repository}, arXiv:2001.07676, 2020.

\bibitem{jiang-etal-2020-know}
Zhengbao Jiang, Frank~F. Xu, Jun Araki, and Graham Neubig.
\newblock How can we know what language models know?
\newblock {\em TACL}, 8:423--438, 2020.

\bibitem{gao2020making}
Tianyu Gao, Adam Fisch, and Danqi Chen.
\newblock Making pre-trained language models better few-shot learners.
\newblock {\em arXiv preprint arXiv:2012.15723}, 2020.

\bibitem{zhao2021calibrate}
Tony~Z. Zhao, Eric Wallace, Shi Feng, Dan Klein, and Sameer Singh.
\newblock Calibrate before use: Improving few-shot performance of language
  models, 2021.

\bibitem{lu2021fantastically}
Yao Lu, Max Bartolo, Alastair Moore, Sebastian Riedel, and Pontus Stenetorp.
\newblock Fantastically ordered prompts and where to find them: Overcoming
  few-shot prompt order sensitivity, 2021.

\bibitem{liu2021what}
Jiachang Liu, Dinghan Shen, Yizhe Zhang, Bill Dolan, L.~Carin, and W.~Chen.
\newblock What makes good in-context examples for gpt-3?
\newblock {\em arXiv}, abs/2101.06804, 2021.

\bibitem{schick2020small}
Timo Schick and Hinrich Schütze.
\newblock It's not just size that matters: Small language models are also
  few-shot learners.
\newblock {\em Computing Research Repository}, arXiv:2009.07118, 2020.

\bibitem{perez2021rissanen}
Ethan Perez, Douwe Kiela, and Kyunghyun Cho.
\newblock Rissanen data analysis: Examining dataset characteristics via
  description length.
\newblock In {\em ICML}, 2021.

\bibitem{schick2020few}
Timo Schick and H.~Schutze.
\newblock Few-shot text generation with pattern-exploiting training.
\newblock {\em arXiv}, abs/2012.11926, 2020.

\bibitem{tam2021improving}
Derek Tam, Rakesh~R Menon, Mohit Bansal, Shashank Srivastava, and Colin Raffel.
\newblock Improving and simplifying pattern exploiting training.
\newblock {\em arxiv preprint arXiv:2103.11955}, 2021.

\bibitem{radford2021learning}
Alec Radford, Jong~Wook Kim, Chris Hallacy, Aditya Ramesh, Gabriel Goh,
  Sandhini Agarwal, Girish Sastry, Amanda Askell, Pamela Mishkin, Jack Clark,
  Gretchen Krueger, and Ilya Sutskever.
\newblock Learning transferable visual models from natural language
  supervision, 2021.

\bibitem{wang2021entailment}
Sinong Wang, Han Fang, Madian Khabsa, Hanzi Mao, and Hao Ma.
\newblock Entailment as few-shot learner, 2021.

\bibitem{scao2021data}
Teven~Le Scao and Alexander~M. Rush.
\newblock How many data points is a prompt worth?, 2021.

\bibitem{sanh2019distilbert}
Victor Sanh, Lysandre Debut, Julien Chaumond, and Thomas Wolf.
\newblock Distilbert, a distilled version of bert: smaller, faster, cheaper and
  lighter.
\newblock {\em arXiv}, abs/1910.01108, 2019.

\bibitem{petroni-etal-2019-language}
Fabio Petroni, Tim Rockt{\"a}schel, Sebastian Riedel, Patrick Lewis, Anton
  Bakhtin, Yuxiang Wu, and Alexander Miller.
\newblock Language models as knowledge bases?
\newblock In {\em EMNLP}, pages 2463--2473, Hong Kong, China, November 2019.
  ACL.

\bibitem{vinyals2016matching}
Oriol Vinyals, Charles Blundell, Timothy Lillicrap, koray kavukcuoglu, and Daan
  Wierstra.
\newblock Matching networks for one shot learning.
\newblock In D.~Lee, M.~Sugiyama, U.~Luxburg, I.~Guyon, and R.~Garnett,
  editors, {\em NeuRIPS}, volume~29. Curran Associates, Inc., 2016.

\bibitem{snell2017prototypical}
Jake Snell, Kevin Swersky, and Richard Zemel.
\newblock Prototypical networks for few-shot learning.
\newblock In I.~Guyon, U.~V. Luxburg, S.~Bengio, H.~Wallach, R.~Fergus,
  S.~Vishwanathan, and R.~Garnett, editors, {\em NeuRIPS}, volume~30. Curran
  Associates, Inc., 2017.

\bibitem{ravi2017optimization}
S.~Ravi and H.~Larochelle.
\newblock Optimization as a model for few-shot learning.
\newblock In {\em ICLR}, 2017.

\bibitem{li2017learning}
Ke~Li and Jitendra Malik.
\newblock Learning to optimize.
\newblock {\em arXiv}, abs/1606.01885, 2017.

\bibitem{allend1974relationship}
David~M. Allen.
\newblock The relationship between variable selection and data agumentation and
  a method for prediction.
\newblock {\em Technometrics}, 16(1):125--127, 1974.

\bibitem{stone1974cross}
M.~Stone.
\newblock Cross‐validatory choice and assessment of statistical predictions.
\newblock {\em Journal of the Royal Statistical Society. Series A
  (Methodological)}, 36:111--133, 1974.

\bibitem{geisser1975predictive}
Seymour Geisser.
\newblock The predictive sample reuse method with applications.
\newblock {\em Journal of the American Statistical Association},
  70(350):320--328, 1975.

\bibitem{hastie2001statistical}
Trevor Hastie, Robert Tibshirani, and Jerome Friedman.
\newblock {\em The Elements of Statistical Learning}.
\newblock Springer Series in Statistics. Springer New York Inc., New York, NY,
  USA, 2001.

\bibitem{finn2017model}
Chelsea Finn, Pieter Abbeel, and Sergey Levine.
\newblock Model-agnostic meta-learning for fast adaptation of deep networks.
\newblock In {\em ICML}, volume~70 of {\em ICML’17}, pages 1126--1135.
  JMLR.org, 2017.

\bibitem{rajeswaran2019meta}
Aravind Rajeswaran, Chelsea Finn, Sham~M Kakade, and Sergey Levine.
\newblock Meta-learning with implicit gradients.
\newblock In H.~Wallach, H.~Larochelle, A.~Beygelzimer, F.~d\textquotesingle
  Alch\'{e}-Buc, E.~Fox, and R.~Garnett, editors, {\em NeurIPS}, volume~32.
  Curran Associates, Inc., 2019.

\bibitem{rissanen1978modeling}
J.~Rissanen.
\newblock Modeling by shortest data description.
\newblock {\em Automatica}, 14(5):465 -- 471, 1978.

\bibitem{rissanen1984universal}
J.~{Rissanen}.
\newblock Universal coding, information, prediction, and estimation.
\newblock {\em IEEE Transactions on Information Theory}, 30(4):629--636, 1984.

\bibitem{dawid1984present}
A.~P. Dawid.
\newblock Present position and potential developments: Some personal views:
  Statistical theory: The prequential approach.
\newblock {\em Journal of the Royal Statistical Society. Series A (General)},
  147(2):278--292, 1984.

\bibitem{grunwald2004tutorial}
Peter Grünwald.
\newblock A tutorial introduction to the minimum description length principle.
\newblock {\em CoRR}, math.ST/0406077, 06 2004.

\bibitem{blier2018description}
L\'{e}onard Blier and Yann Ollivier.
\newblock The description length of deep learning models.
\newblock In S.~Bengio, H.~Wallach, H.~Larochelle, K.~Grauman, N.~Cesa-Bianchi,
  and R.~Garnett, editors, {\em NeuRIPS}, volume~31, pages 2216--2226. Curran
  Associates, Inc., 2018.

\bibitem{blumer1987occam}
Alselm Blumer, Andrzej Ehrenfeucht, David Haussler, and Manfred~K. Warmuth.
\newblock Occam's razor.
\newblock {\em Inf. Process. Lett.}, 24(6):377–380, April 1987.

\bibitem{sinha2021masked}
Koustuv Sinha, Robin Jia, Dieuwke Hupkes, Joelle Pineau, Adina Williams, and
  Douwe Kiela.
\newblock Masked language modeling and the distributional hypothesis: Order
  word matters pre-training for little.
\newblock {\em CoRR}, abs/2104.06644, 2021.

\bibitem{phillips2000feret}
P.~J. {Phillips}, {Hyeonjoon Moon}, S.~A. {Rizvi}, and P.~J. {Rauss}.
\newblock The feret evaluation methodology for face-recognition algorithms.
\newblock {\em TPAMI}, 22(10):1090--1104, 2000.

\bibitem{buolamwini2018gender}
Joy Buolamwini and Timnit Gebru.
\newblock Gender shades: Intersectional accuracy disparities in commercial
  gender classification.
\newblock In Sorelle~A. Friedler and Christo Wilson, editors, {\em Fairness,
  Accountability and Transparency}, volume~81 of {\em PMLR}, pages 77--91, New
  York, NY, USA, 23--24 Feb 2018. PMLR.

\bibitem{henderson2018ethical}
Peter Henderson, Koustuv Sinha, Nicolas Angelard-Gontier, Nan~Rosemary Ke,
  Genevieve Fried, Ryan Lowe, and Joelle Pineau.
\newblock Ethical challenges in data-driven dialogue systems.
\newblock In {\em AAAI/ACM Conference on AI, Ethics, and Society}, AIES '18,
  page 123–129, New York, NY, USA, 2018. Association for Computing Machinery.

\bibitem{khatri2018advancing}
Chandra Khatri, Behnam Hedayatnia, Anu Venkatesh, Jeff Nunn, Yi~Pan, Qing Liu,
  Han Song, Anna Gottardi, Sanjeev Kwatra, Sanju Pancholi, Ming Cheng, Qinglang
  Chen, Lauren Stubel, Karthik Gopalakrishnan, Kate Bland, Raefer Gabriel,
  Arindam Mandal, Dilek Hakkani{-}T{\"{u}}r, Gene Hwang, Nate Michel, Eric
  King, and Rohit Prasad.
\newblock Advancing the state of the art in open domain dialog systems through
  the alexa prize.
\newblock {\em CoRR}, abs/1812.10757, 2018.

\bibitem{garcia2015comprehensive}
Javier Garc{{\'i}}a, Fern, and o~Fern{{\'a}}ndez.
\newblock A comprehensive survey on safe reinforcement learning.
\newblock {\em JMLR}, 16(42):1437--1480, 2015.

\bibitem{amodei2016concrete}
Dario Amodei, Chris Olah, Jacob Steinhardt, Paul~F. Christiano, John Schulman,
  and Dan Man{\'{e}}.
\newblock Concrete problems in {AI} safety.
\newblock {\em CoRR}, abs/1606.06565, 2016.

\bibitem{watanabe2010asymptotic}
Sumio Watanabe.
\newblock Asymptotic equivalence of bayes cross validation and widely
  applicable information criterion in singular learning theory.
\newblock {\em JMLR}, 11(116):3571--3594, 2010.

\bibitem{akaike1974new}
H.~Akaike.
\newblock A new look at the statistical model identification.
\newblock {\em TACON}, 19(6):716--723, 1974.

\bibitem{mallows1973some}
C.~L. Mallows.
\newblock Some comments on cp.
\newblock {\em Technometrics}, 15(4):661--675, 1973.

\bibitem{hutter2011sequential}
Frank Hutter, Holger~H. Hoos, and Kevin Leyton-Brown.
\newblock Sequential model-based optimization for general algorithm
  configuration.
\newblock In Carlos A.~Coello Coello, editor, {\em Learning and Intelligent
  Optimization}, pages 507--523, Berlin, Heidelberg, 2011. Springer Berlin
  Heidelberg.

\bibitem{bergstra2011algorithms}
James Bergstra, R\'{e}mi Bardenet, Yoshua Bengio, and Bal\'{a}zs K\'{e}gl.
\newblock Algorithms for hyper-parameter optimization.
\newblock In J.~Shawe-Taylor, R.~Zemel, P.~Bartlett, F.~Pereira, and K.~Q.
  Weinberger, editors, {\em NeuRIPS}, volume~24. Curran Associates, Inc., 2011.

\bibitem{snoek2012practical}
Jasper Snoek, Hugo Larochelle, and Ryan~P Adams.
\newblock Practical bayesian optimization of machine learning algorithms.
\newblock In F.~Pereira, C.~J.~C. Burges, L.~Bottou, and K.~Q. Weinberger,
  editors, {\em NeuRIPS}, volume~25. Curran Associates, Inc., 2012.

\bibitem{miikkulainen2019evolving}
Risto Miikkulainen, Jason Liang, Elliot Meyerson, Aditya Rawal, Daniel Fink,
  Olivier Francon, Bala Raju, Hormoz Shahrzad, Arshak Navruzyan, Nigel Duffy,
  and Babak Hodjat.
\newblock Chapter 15 - evolving deep neural networks.
\newblock In Robert Kozma, Cesare Alippi, Yoonsuck Choe, and Francesco~Carlo
  Morabito, editors, {\em Artificial Intelligence in the Age of Neural Networks
  and Brain Computing}, pages 293--312. Academic Press, 2019.

\bibitem{real2019regularized}
Esteban Real, Alok Aggarwal, Yanping Huang, and Quoc~V. Le.
\newblock Regularized evolution for image classifier architecture search.
\newblock {\em AAAI}, 33(01):4780--4789, Jul. 2019.

\bibitem{zoph2017neural}
Barret Zoph and Quoc~V. Le.
\newblock Neural architecture search with reinforcement learning.
\newblock In {\em ICLR}. OpenReview.net, 2017.

\bibitem{larsen1996design}
J.~Larsen, L.K. Hansen, C.~Svarer, and M.~Ohlsson.
\newblock Design and regularization of neural networks: the optimal use of a
  validation set.
\newblock In {\em Neural Networks for Signal Processing VI. IEEE Signal
  Processing Society Workshop}, pages 62--71, 1996.

\bibitem{bengio2000gradient}
Yoshua Bengio.
\newblock {Gradient-Based Optimization of Hyperparameters}.
\newblock {\em Neural Computation}, 12(8):1889--1900, 08 2000.

\bibitem{chapelle2004choosing}
O.~Chapelle, V.~Vapnik, O.~Bousquet, and S.~Mukherjee.
\newblock Choosing multiple parameters for support vector machines.
\newblock {\em Machine Learning}, 46:131--159, 2004.

\bibitem{liu2018darts}
Hanxiao Liu, Karen Simonyan, and Yiming Yang.
\newblock {DARTS}: Differentiable architecture search.
\newblock In {\em ICLR}, 2019.

\bibitem{shin-etal-2020-autoprompt}
Taylor Shin, Yasaman Razeghi, Robert~L. Logan~IV, Eric Wallace, and Sameer
  Singh.
\newblock {A}uto{P}rompt: {E}liciting {K}nowledge from {L}anguage {M}odels with
  {A}utomatically {G}enerated {P}rompts.
\newblock In {\em EMNLP}, pages 4222--4235, Online, November 2020. ACL.

\bibitem{liu2021gpt}
Xiao Liu, Yanan Zheng, Zhengxiao Du, Ming Ding, Yujie Qian, Zhilin Yang, and
  Jie Tang.
\newblock Gpt understands, too, 2021.

\bibitem{zhong2021factual}
Zexuan Zhong, Dan Friedman, and Danqi Chen.
\newblock Factual probing is {[MASK]:} learning vs. learning to recall.
\newblock {\em CoRR}, abs/2104.05240, 2021.

\bibitem{poerner-etal-2020-e}
Nina Poerner, Ulli Waltinger, and Hinrich Sch{\"u}tze.
\newblock {E}-{BERT}: Efficient-yet-effective entity embeddings for {BERT}.
\newblock In {\em Findings of EMNLP}, pages 803--818, Online, November 2020.
  ACL.

\bibitem{wolf-etal-2020-transformers}
Thomas Wolf, Lysandre Debut, Victor Sanh, Julien Chaumond, Clement Delangue,
  Anthony Moi, Pierric Cistac, Tim Rault, Rémi Louf, Morgan Funtowicz, Joe
  Davison, Sam Shleifer, Patrick von Platen, Clara Ma, Yacine Jernite, Julien
  Plu, Canwen Xu, Teven~Le Scao, Sylvain Gugger, Mariama Drame, Quentin Lhoest,
  and Alexander~M. Rush.
\newblock Transformers: State-of-the-art natural language processing.
\newblock In {\em EMNLP: System Demonstrations}, pages 38--45, Online, October
  2020. ACL.

\bibitem{paszke2019pytorch}
Adam Paszke, Sam Gross, Francisco Massa, Adam Lerer, James Bradbury, Gregory
  Chanan, Trevor Killeen, Zeming Lin, Natalia Gimelshein, Luca Antiga, Alban
  Desmaison, Andreas Kopf, Edward Yang, Zachary DeVito, Martin Raison, Alykhan
  Tejani, Sasank Chilamkurthy, Benoit Steiner, Lu~Fang, Junjie Bai, and Soumith
  Chintala.
\newblock Pytorch: An imperative style, high-performance deep learning library.
\newblock In H.~Wallach, H.~Larochelle, A.~Beygelzimer, F.~d\textquotesingle
  Alch\'{e}-Buc, E.~Fox, and R.~Garnett, editors, {\em NeuRIPS}, pages
  8024--8035. Curran Associates, Inc., 2019.

\bibitem{voita-titov-2020-information}
Elena Voita and Ivan Titov.
\newblock Information-theoretic probing with minimum description length.
\newblock In {\em EMNLP}, pages 183--196, Online, November 2020. ACL.

\bibitem{wang2019superglue}
Alex Wang, Yada Pruksachatkun, Nikita Nangia, Amanpreet Singh, Julian Michael,
  Felix Hill, Omer Levy, and Samuel Bowman.
\newblock Superglue: A stickier benchmark for general-purpose language
  understanding systems.
\newblock In H.~Wallach, H.~Larochelle, A.~Beygelzimer, F.~d\textquotesingle
  Alch\'{e}-Buc, E.~Fox, and R.~Garnett, editors, {\em NeuRIPS}, volume~32.
  Curran Associates, Inc., 2019.

\bibitem{lan2020albert}
Zhenzhong Lan, Mingda Chen, Sebastian Goodman, Kevin Gimpel, Piyush Sharma, and
  Radu Soricut.
\newblock Albert: A lite bert for self-supervised learning of language
  representations.
\newblock In {\em ICLR}, 2020.

\bibitem{devlin-etal-2019-bert}
Jacob Devlin, Ming-Wei Chang, Kenton Lee, and Kristina Toutanova.
\newblock {BERT}: Pre-training of deep bidirectional transformers for language
  understanding.
\newblock In {\em NAACL}, pages 4171--4186, Minneapolis, Minnesota, June 2019.
  ACL.

\bibitem{triantafillou2020metadataset}
Eleni Triantafillou, Tyler Zhu, Vincent Dumoulin, Pascal Lamblin, Utku Evci,
  Kelvin Xu, Ross Goroshin, Carles Gelada, Kevin Swersky, Pierre-Antoine
  Manzagol, and Hugo Larochelle.
\newblock Meta-dataset: A dataset of datasets for learning to learn from few
  examples.
\newblock In {\em ICLR}, 2020.

\bibitem{ye2021crossfit}
Qinyuan Ye, Bill~Yuchen Lin, and Xiang Ren.
\newblock Crossfit: {A} few-shot learning challenge for cross-task
  generalization in {NLP}.
\newblock {\em CoRR}, abs/2104.08835, 2021.

\bibitem{caruana1995learning}
Rich Caruana.
\newblock Learning many related tasks at the same time with backpropagation.
\newblock In G.~Tesauro, D.~Touretzky, and T.~Leen, editors, {\em NeuRIPS},
  volume~7. MIT Press, 1995.

\bibitem{caruana1997multitask}
Rich Caruana.
\newblock Multitask learning.
\newblock {\em Machine Learning}, 28(1):41–75, July 1997.

\bibitem{phang2018stilts}
Jason Phang, Thibault F\'evry, and Samuel~R. Bowman.
\newblock Sentence encoders on stilts: Supplementary training on intermediate
  labeled-data tasks.
\newblock {\em CoRR}, abs/1811.01088, 2018.

\bibitem{liu-etal-2019-multi}
Xiaodong Liu, Pengcheng He, Weizhu Chen, and Jianfeng Gao.
\newblock Multi-task deep neural networks for natural language understanding.
\newblock In {\em ACL}, pages 4487--4496, Florence, Italy, July 2019. ACL.

\bibitem{kocijan-etal-2019-surprisingly}
Vid Kocijan, Ana-Maria Cretu, Oana-Maria Camburu, Yordan Yordanov, and Thomas
  Lukasiewicz.
\newblock A surprisingly robust trick for the {W}inograd schema challenge.
\newblock In {\em ACL}, pages 4837--4842, Florence, Italy, July 2019. ACL.

\bibitem{xie2020unsupervised}
Qizhe Xie, Zihang Dai, Eduard Hovy, Thang Luong, and Quoc Le.
\newblock Unsupervised data augmentation for consistency training.
\newblock In H.~Larochelle, M.~Ranzato, R.~Hadsell, M.~F. Balcan, and H.~Lin,
  editors, {\em NeuRIPS}, volume~33, pages 6256--6268. Curran Associates, Inc.,
  2020.

\bibitem{chen-etal-2020-mixtext}
Jiaao Chen, Zichao Yang, and Diyi Yang.
\newblock {M}ix{T}ext: Linguistically-informed interpolation of hidden space
  for semi-supervised text classification.
\newblock In {\em ACL}, pages 2147--2157, Online, July 2020. ACL.

\bibitem{yang-etal-2020-generative}
Yiben Yang, Chaitanya Malaviya, Jared Fernandez, Swabha Swayamdipta, Ronan
  Le~Bras, Ji-Ping Wang, Chandra Bhagavatula, Yejin Choi, and Doug Downey.
\newblock Generative data augmentation for commonsense reasoning.
\newblock In {\em Findings of EMNLP}, pages 1008--1025, Online, November 2020.
  ACL.

\bibitem{artetxe2018unsupervised-neural}
Mikel Artetxe, Gorka Labaka, Eneko Agirre, and Kyunghyun Cho.
\newblock Unsupervised neural machine translation.
\newblock In {\em ICLR}, 2018.

\bibitem{lample2018unsupervised}
Guillaume Lample, Alexis Conneau, Ludovic Denoyer, and Marc'Aurelio Ranzato.
\newblock Unsupervised machine translation using monolingual corpora only.
\newblock In {\em ICLR}, 2018.

\bibitem{perez-etal-2020-unsupervised}
Ethan Perez, Patrick Lewis, Wen-tau Yih, Kyunghyun Cho, and Douwe Kiela.
\newblock Unsupervised question decomposition for question answering.
\newblock In {\em EMNLP}, pages 8864--8880, Online, November 2020. ACL.

\bibitem{oliver2018realistic}
Avital Oliver, Augustus Odena, Colin Raffel, Ekin~D. Cubuk, and Ian~J.
  Goodfellow.
\newblock Realistic evaluation of deep semi-supervised learning algorithms.
\newblock {\em CoRR}, abs/1804.09170, 2018.

\bibitem{dodge-etal-2019-show}
Jesse Dodge, Suchin Gururangan, Dallas Card, Roy Schwartz, and Noah~A. Smith.
\newblock Show your work: Improved reporting of experimental results.
\newblock In {\em EMNLP}, pages 2185--2194, Hong Kong, China, November 2019.
  ACL.

\bibitem{ettinger-2020-bert}
Allyson Ettinger.
\newblock What {BERT} is not: Lessons from a new suite of psycholinguistic
  diagnostics for language models.
\newblock {\em TACL}, 8:34--48, 2020.

\bibitem{vehtari2017practical}
Aki Vehtari, Andrew Gelman, and Jonah Gabry.
\newblock Practical bayesian model evaluation using leave-one-out
  cross-validation and waic.
\newblock {\em Statistics and Computing}, 27(5):1413–1432, September 2017.

\bibitem{dodge2020finetuning}
Jesse Dodge, Gabriel Ilharco, Roy Schwartz, Ali Farhadi, Hannaneh Hajishirzi,
  and Noah~A. Smith.
\newblock Fine-tuning pretrained language models: Weight initializations, data
  orders, and early stopping.
\newblock {\em CoRR}, abs/2002.06305, 2020.

\bibitem{clark-etal-2019-boolq}
Christopher Clark, Kenton Lee, Ming-Wei Chang, Tom Kwiatkowski, Michael
  Collins, and Kristina Toutanova.
\newblock {B}ool{Q}: Exploring the surprising difficulty of natural yes/no
  questions.
\newblock In {\em NAACL}, pages 2924--2936, Minneapolis, Minnesota, June 2019.
  ACL.

\bibitem{de2019commitment}
Marie-Catherine de~Marneffe, Mandy Simons, and Judith Tonhauser.
\newblock The commitmentbank: Investigating projection in naturally occurring
  discourse.
\newblock {\em Sinn und Bedeutung}, 23(2):107--124, July 2019.

\bibitem{dagan2006pascal}
Ido Dagan, Oren Glickman, and Bernardo Magnini.
\newblock The pascal recognising textual entailment challenge.
\newblock In Joaquin Qui{\~{n}}onero-Candela, Ido Dagan, Bernardo Magnini, and
  Florence d'Alch{\'e} Buc, editors, {\em Machine Learning Challenges.
  Evaluating Predictive Uncertainty, Visual Object Classification, and
  Recognising Tectual Entailment}, pages 177--190, Berlin, Heidelberg, 2006.
  Springer Berlin Heidelberg.

\bibitem{bar2006second}
Roy Bar~Haim, Ido Dagan, Bill Dolan, Lisa Ferro, Danilo Giampiccolo, Bernardo
  Magnini, and Idan Szpektor.
\newblock The second {PASCAL} recognising textual entailment challenge.
\newblock In {\em Second {PASCAL} Challenges Workshop on Recognising Textual
  Entailment}, 2006.

\bibitem{giampiccolo2007pascal}
Danilo Giampiccolo, Bernardo Magnini, Ido Dagan, and Bill Dolan.
\newblock The third pascal recognizing textual entailment challenge.
\newblock In {\em ACL-PASCAL Workshop on Textual Entailment and Paraphrasing},
  RTE '07, page 1–9, USA, 2007. ACL.

\bibitem{bentivogli2009fifth}
Luisa Bentivogli, Ido Dagan, Hoa~Trang Dang, Danilo Giampiccolo, and Bernardo
  Magnini.
\newblock The fifth pascal recognizing textual entailment challenge.
\newblock In {\em TAC}, 2009.

\bibitem{pilehvar2018wic}
Mohammad~Taher Pilehvar and Jose Camacho-Collados.
\newblock {WiC}: The word-in-context dataset for evaluating context-sensitive
  meaning representations.
\newblock In {\em NAACL}. ACL, 2019.

\bibitem{levesque2012winograd}
Hector~J. Levesque, Ernest Davis, and Leora Morgenstern.
\newblock The winograd schema challenge.
\newblock In {\em KR}, KR'12, page 552–561. AAAI Press, 2012.

\bibitem{khashabi2018looking}
Daniel Khashabi, Snigdha Chaturvedi, Michael Roth, Shyam Upadhyay, and Dan
  Roth.
\newblock Looking beyond the surface: A challenge set for reading comprehension
  over multiple sentences.
\newblock In {\em NAACL}. ACL, 2018.

\bibitem{zhang2018record}
Sheng Zhang, Xiaodong Liu, Jingjing Liu, Jianfeng Gao, Kevin Duh, and Benjamin
  Van~Durme.
\newblock {R}e{C}o{RD}: Bridging the gap between human and machine commonsense
  reading comprehension.
\newblock {\em arXiv preprint 1810.12885}, 2018.

\end{thebibliography}

\clearpage
\appendix

\section{True Few-Shot Prompt Selection with Other Generalization Criteria}
\label{sec:True Few-Shot Prompt Selection with Other Generalization Criteria}
Here, we evaluate the performance of prompts chosen using other generalization criteria, to examine the extent to which poor prompt selection is specific to CV and MDL.
We evaluate on LAMA and follow the same experimental setup used to evaluate CV/MDL, as described in \S\ref{ssec:Experimental Setup}.
As before, we examine the average test accuracy of the prompt chosen by a particular criterion, as well as the percentage of the time that a given criterion chose the prompt with the highest test accuracy.
We now describe the other criteria we test.

\subsection{Bayesian Cross-Validation}

Bayesian CV is a variant of CV that evaluates a learning algorithm $\mathcal{A}$ based on its expected loss on a held-out fold after marginalizing over the model according the posterior distribution~\citep[for an overview, see][]{vehtari2017practical}.
In our setup, each model corresponds to a unique set of random factors $R$ trained by $\mathcal{A}$.
Given some inputs $X=x_{1:N}$ and labels $Y=y_{1:N}$, we assume a uniform prior $p(R)$ over $R$ and assume that $R$ and $X$ are independent ($p(R|X)=p(R)$). We then derive the posterior probability as:
\begin{align*}
    p(R|X, Y) = \frac{p(Y|R, X) p(R | X)}{p(Y | X)} = \frac{p(Y | R, X)}{p(Y | X)}=\frac{p(Y|R, X)}{\sum_{R'} p(Y|R', X)}
\end{align*}
where for any $R'$:
\begin{align*}
    p(Y | R', X) = \prod_{i=1}^{N} p(y_{i} | y_{1:i-1}, X, R') = \prod_{i=1}^{N} p(y_{i} | y_{1:i-1}, x_{1:i}, R').
\end{align*}
The second equality holds because $p$ is a left-to-right LM that predicts $y_i$ only based on the input $x_i$ and earlier examples $(x_{1:i-1}, y_{1:i-1})$.
We marginalize out the model over the posterior distribution:
\begin{align*}
    \text{CV}_{\text{Bayes}}(\mathcal{A},R,F) = \mathbb{E}_{k \sim \text{Unif}(1, K)} \bigg[ \mathcal{L} \Big( \mathbb{E}_{R \sim p(R | F(D_{\text{train}})_{\neg k})}[\mathcal{A}(F(D_{\text{train}})_{\neg k}, R)]; F(D_{\text{train}})_{k} \Big) \bigg]
\end{align*}
We then choose the algorithm (prompt) that minimizes $\mathbb{E}_{R,F}[\text{CV}_{\text{Bayes}}(\mathcal{A},R,F)]$, where $R$ is the order of training examples.

\subsection{Interpolating between CV and MDL}
Our experiments in the main paper suggest that CV/MDL behave differently in terms of prompt selection.
In this section, we describe a way to interpolate between CV and MDL, in order to devise a new criterion that may inherit advantageous properties from both CV and MDL.
Similar to MDL, we measure the expected loss on a held-out fold $F(D_{\text{train}})_k$  when training on the previous $F(D_{\text{train}})_{1:k-1}$ folds, doing so across all $k=1, \dots, K$ folds.
However, we now weight the loss on $F(D_{\text{train}})_k$ by a factor that depends on the number of training examples, $p(k; \beta) \propto \text{exp}(-\beta |F(D_{\text{train}})_{1:k-1}|)$, where $\beta$ is an inverse temperature hyperparameter.
MDL is equivalent to using a uniform weight over all train sizes ($\beta=0$), and CV is equivalent to using a non-zero weight for only the largest train size ($\beta=\infty$).
Formally, we define the interpolated criteria, $\text{MDL}_{\beta}(\mathcal{A}, R, F)$, as follows:
\begin{align*}
    \text{MDL}_{\beta}(\mathcal{A},R,F) = \mathbb{E}_{k \sim p(k; \beta)} \bigg[ \mathcal{L} \Big( \mathcal{A}(F(D_{\text{train}})_{1:k-1}, R); F(D_{\text{train}})_{k} \Big) \bigg].
\end{align*}
We set the hyperparameter $\beta$ to the default value of $\beta=1$ to avoid having to choose $\beta$ based on a limited number of examples available in true few-shot learning.
We choose the algorithm that minimizes $\mathbb{E}_{R,F}[\text{MDL}_{\beta}(\mathcal{A},R,F)]$.

\subsection{Joint Log-Probability}
Up to this point, we have used generalization criteria that use the NLL of the label given the input, $-\log p(y|x)$, as the loss function $\mathcal{L}$.
However, other loss functions may correlate better with generalization.
In particular, we hypothesize that a good prompt leads the LM to give the entire input $(x, y)$ high probability, i.e., a low, joint log-probability $-\log p(x, y)$.
We thus use $-\log p(x, y)$ as the loss function to measure CV and MDL, which we refer to as $\text{CV}_{x, y}$ and $\text{MDL}_{x, y}$, respectively.
Since $-\log p(x, y) = [-\log p(y|x)] + [-\log p(x)]$, joint log-probability is equivalent to the label NLL $-\log p(y|x)$ used before, with an additional term $-\log p(x)$ that measures the input NLL.
We measure $-\log p(x, y)$ by evaluating the total NLL of all tokens in the prompt-formatted $(x, y)$ pair (including prompt tokens).
We choose the algorithm that minimizes $\mathbb{E}_{R,F}[\text{CV}_{x,y}(\mathcal{A},R,F)]$ or $\mathbb{E}_{R,F}[\text{MDL}_{x,y}(\mathcal{A},R,F)]$.

\subsection{Results}
\begin{figure*}[t]
	\centering
    \includegraphics[scale=.38]{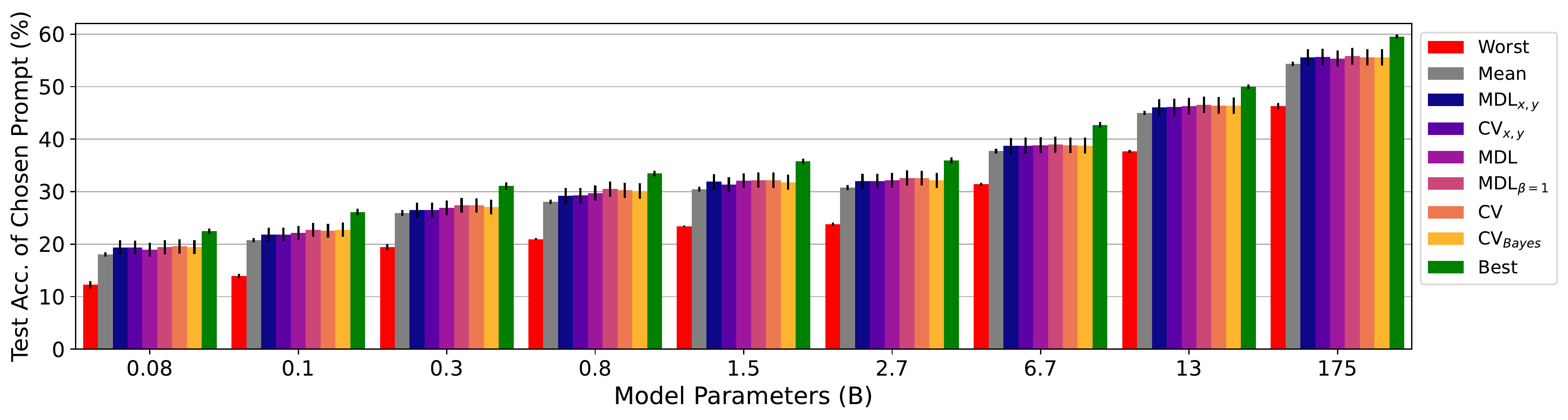}
    \includegraphics[scale=.38]{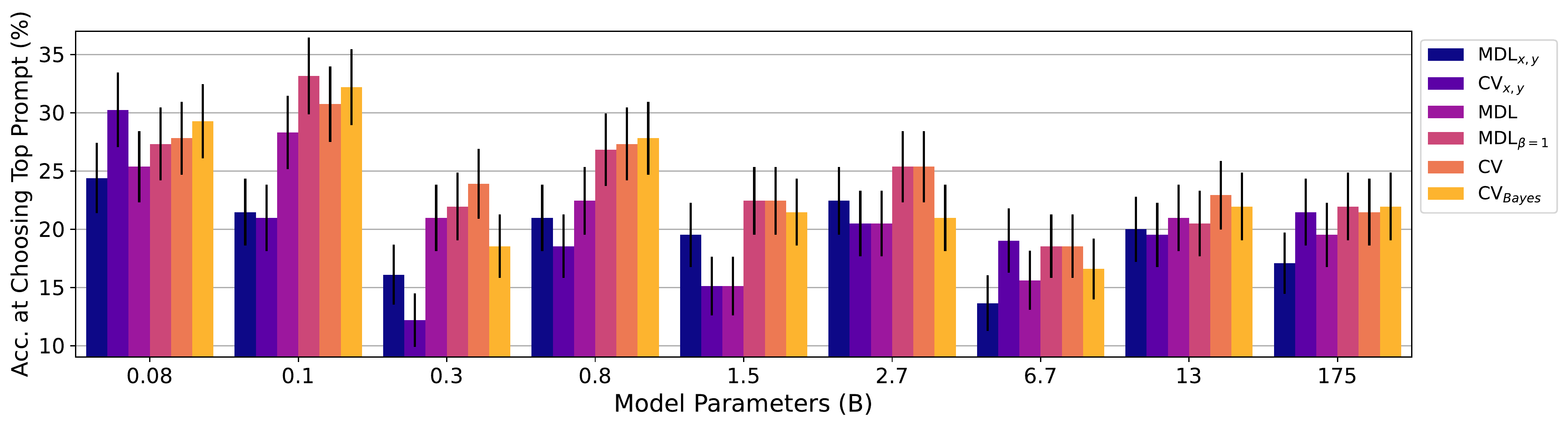}
    \caption{\textbf{Top}: LAMA-UHN accuracy of prompts chosen using different generalization criteria vs. accuracy of the worst, average (randomly-selected), and best prompt (prior work). \textbf{Bottom}: The average accuracy gain from using criteria-chosen prompts instead of randomly-chosen ones, relative to the gain from the best prompt. We plot mean/std. err. across 5 runs with different training sets. Across all model sizes, criteria-chosen prompts obtain only small improvements over randomly-chosen ones and perform far worse than the best prompts.}
    \label{fig:plot_results_by_engine_vary_criterion}
\end{figure*}

As shown in Fig.~\ref{fig:plot_results_by_engine_vary_criterion} (top), all criteria choose prompts with a similar average accuracy, close to the average accuracy of randomly-chosen prompts.
Likewise, all criteria are similarly inaccurate at choosing the highest accuracy prompt, as shown in Fig~\ref{fig:plot_results_by_engine_vary_criterion} (bottom).
These results show that true few-shot prompt selection is challenging not only for CV and MDL but also many other criteria.

\section{Additional Results with MDL}
\label{sec:Additional Results with MDL}

\begin{figure*}[t]
	\centering
        \includegraphics[scale=.28]{images/lama/plot_results_distribution.plot_type-Test-Accuracy.num_train-5.plot_info-cdf.stat-nlls_cv.pdf}
        \includegraphics[scale=.28]{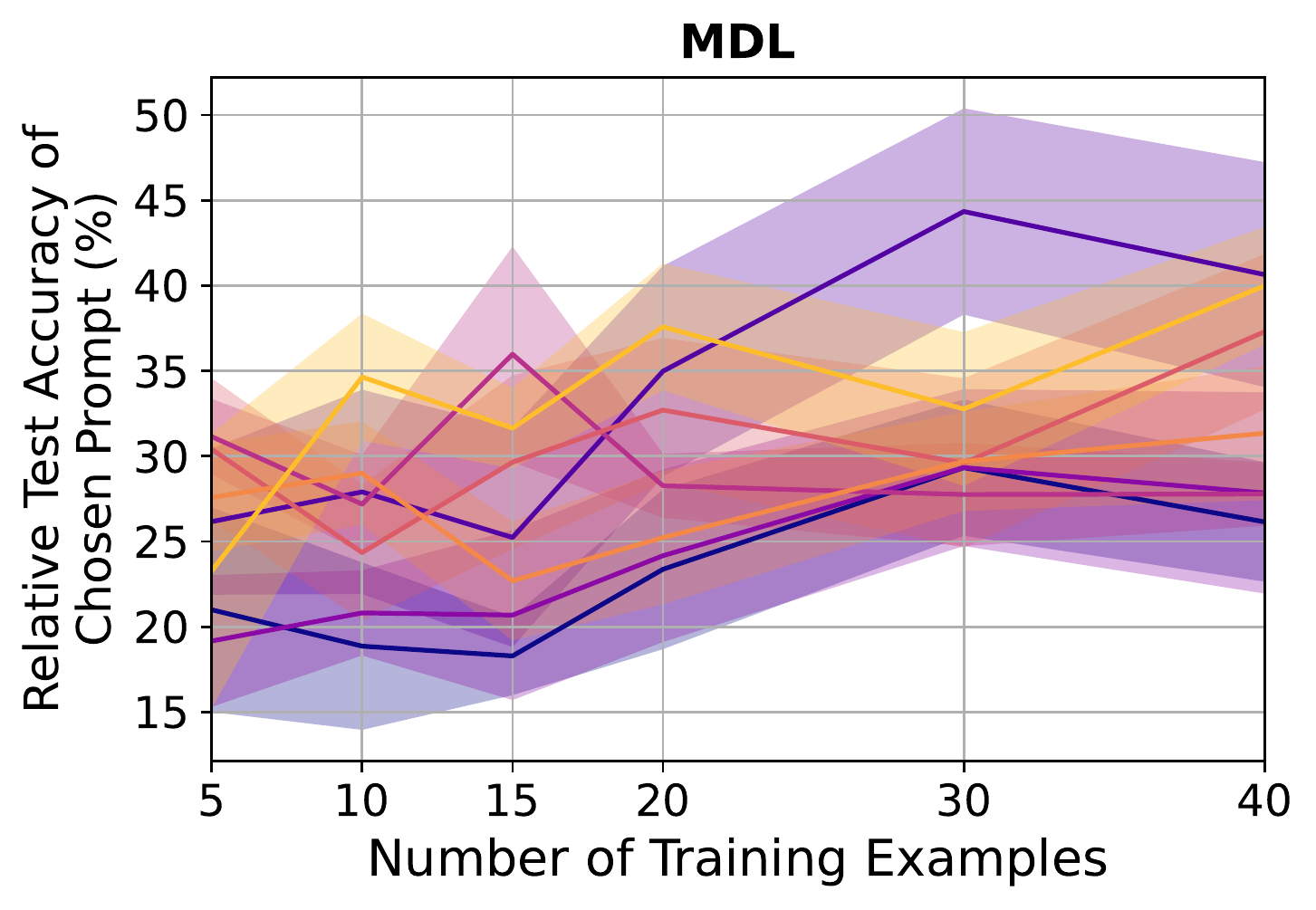}
        \includegraphics[scale=.28]{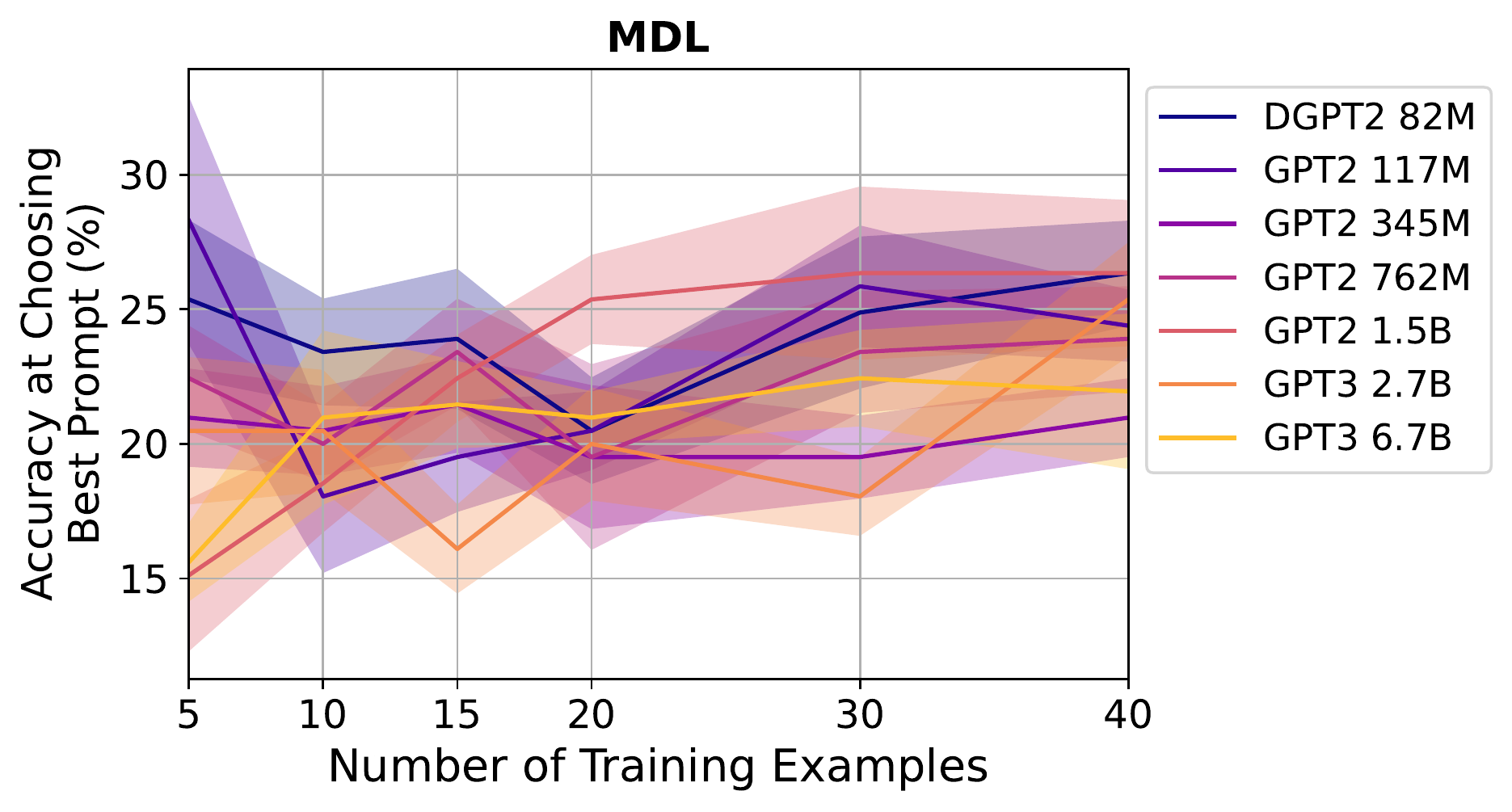}
    \caption{\textbf{Left}: Chance of various accuracy gains for MDL-chosen prompts over average (randomly-chosen) prompts on LAMA-UHN. As with CV, there is a wide variance in accuracy gains, especially for larger models, and a significant chance of choosing a worse-than-average prompt.
    \textbf{Middle}:
    Increasing the number of examples up to 40 does not clearly improve MDL in terms of acc. gain over the average prompt (scaled to 0), relative to the best one (scaled to 100) or (\textbf{Right}) acc. at choosing the best prompt (mean/std. err. on LAMA over 5 runs with different train sets).}
    \label{fig:plot_results_by_num_train-mdl}
\end{figure*}

\begin{figure*}[t]
	\centering
    \includegraphics[scale=.33]{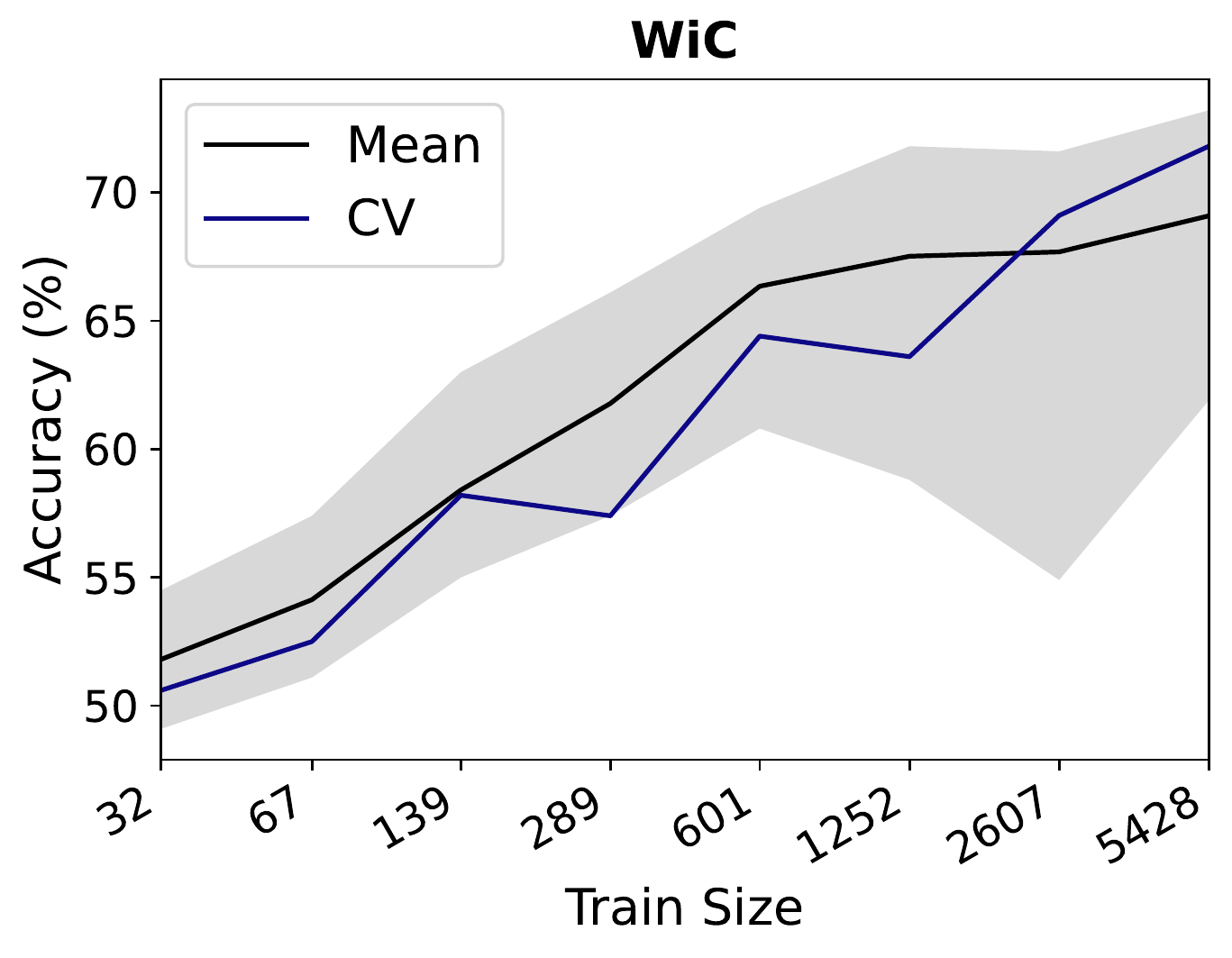}
    \includegraphics[scale=.33]{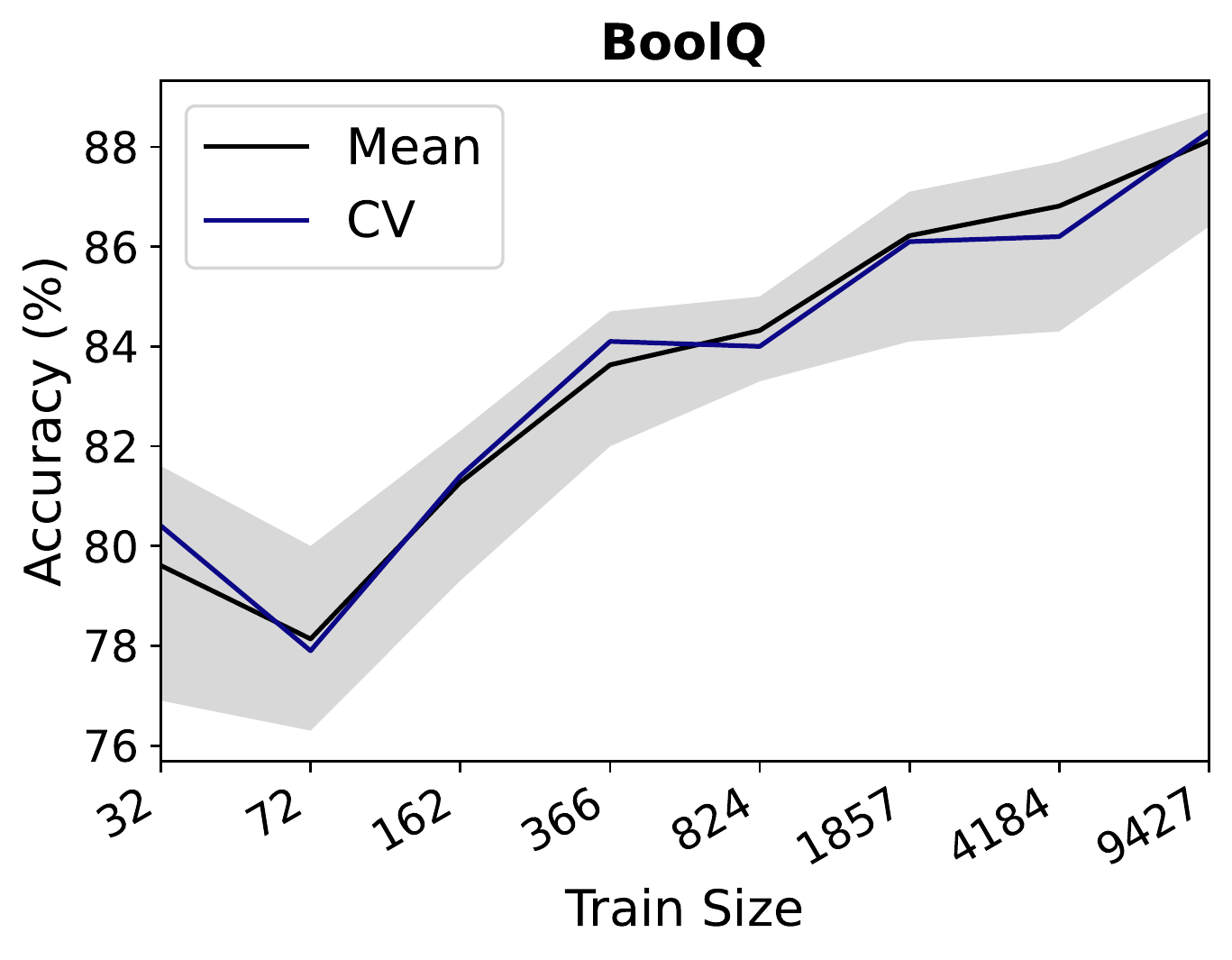}
    \includegraphics[scale=.33]{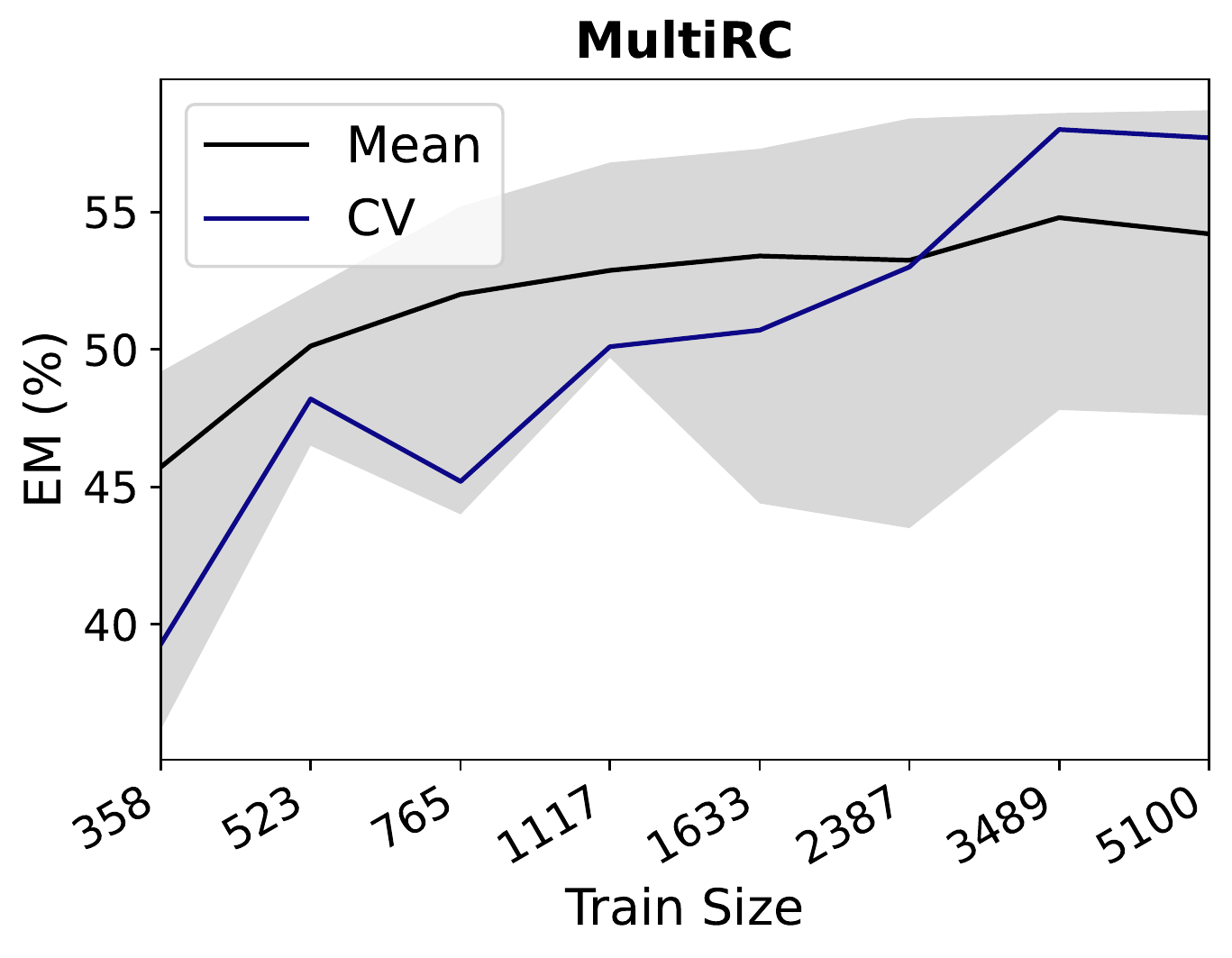}
    \caption{ADAPET accuracy using CV-chosen hyperparameters as the number of examples increases. The shaded region shows the range of accuracies obtained using the same training set but different hyperparameter settings (16 in total).
    }
    \label{fig:plot_results_by_num_train_adapet}
\end{figure*}

In the main paper, we showed several results for CV alone for brevity, so in this section, we show the corresponding plots for MDL as well. The overall trends are the same for both CV and MDL.

In \S\ref{ssec:How reliably does prompt selection improve over the average prompt?}, we found that the gains from choosing prompts using CV are high variance, a variance that increases with model size (Fig.~\ref{fig:plot_results_distribution}).
Here, we show the same results but for MDL in Fig.~\ref{fig:plot_results_by_num_train-mdl} (left).
Similar to CV, MDL-chosen prompts have high variance in test accuracy relative to the average prompt, especially for larger models.
This finding suggests that the high variance is due not to CV in particular, but to the inherent difficulty of true few-shot model selection.

In \S\ref{ssec:Does prompt selection improve with more labeled examples?}, we examined if increasing the number of examples improves prompt selection for CV.
Fig.~\ref{fig:plot_results_by_num_train-mdl} (middle/right) shows the results for MDL, which are similar to those for CV.
When increasing the examples used, we do not observe a consistent increase in the gain achieved by MDL over random selection, relative to the best prompt (Fig.~\ref{fig:plot_results_by_num_train-mdl} middle).
Similarly, we do not observe a consistent increase in the accuracy of MDL at choosing the best prompt (Fig.~\ref{fig:plot_results_by_num_train-mdl} right).
For some model sizes, there may potentially be some improvement with more examples, but the standard error is high, and the overall accuracies achieved by MDL are still lower than those from CV shown earlier in Fig.~\ref{fig:plot_results_by_num_train-loo}.
Overall, model selection is challenging for both CV and MDL, even as we approach the maximum number of examples that can fit in the context of GPT models.

\section{How many examples do you need for effective model selection?}
\label{sec:How many examples do you need for effective model selection?}
Here, we conduct a preliminary analysis on the minimum number of examples is necessary to choose a better-than-average model.
We examine this question in the context of ADAPET, which can handle an arbitrary number of examples (GPT-based models can only handle a number of examples that fit within the LM input---2048 tokens or $\sim$1500 words).
We use the same setup and hyperparameter range as in \S\ref{sec:True Few-Shot Hyperparameter Selection} but vary the number of training examples.

Fig.~\ref{fig:plot_results_by_num_train_adapet} shows accuracy on WiC and BoolQ of CV-chosen hyperparameters, compared to the worst, average, and best hyperparameters.
For WiC and MultiRC, CV requires $>$2-3k examples to choose better-than-average hyperparameters.
For BoolQ, CV performs similar to the average hyperparameters even when using up to 9k examples.
This result may be due to the fact that we retrain the model using the CV-chosen hyperparameters, but finetuning pretrained LMs often has high variance in performance~\cite{phang2018stilts,dodge2020finetuning}.
Thus, when more data is available, CV may be outperformed by using a single train-validation split and choosing the model that does well on the validation split, without retraining on the combined train+validation set.
We leave further exploration of model selection in higher data regimes as an important direction for future work.

\section{Task and Experimental Details}
\label{sec:Task Details}

\subsection{LAMA}
\label{ssec:LAMA}

\paragraph{Prompts Used} For the full list LPAQA prompts, please see \url{https://github.com/jzbjyb/LPAQA/tree/master/prompt}.
There are up to 90 LPAQA prompts per relation, so we use a subset of prompts to evaluate the impact of a small amount of validation-based prompt tuning.
We filter out prompts that do not end with the target answer blanked out (``Geoffrey Hinton was \_ profession.''), which cannot be easily used with left-to-right LMs like GPT.
For mined prompts (group 2), we choose the 5 prompts that occur most frequently in Wikipedia, similar to~\cite{jiang-etal-2020-know}.
We include all prompts if fewer than 5 are available.
For paraphrased prompts (groups 1 and 3), we choose up to 5 prompts with the highest round-trip back-translation probability, similar to~\cite{jiang-etal-2020-know}.
Finally, we de-duplicate prompts, as some prompts occur in multiple groups.

\begin{table*}
\centering
\aboverulesep = 0.205mm
\belowrulesep = 0.405mm
\setlength{\tabcolsep}{2pt}
\begin{tabular}{c|p{11.2cm}|p{2.3cm}}
\toprule
\textbf{Task} & \textbf{Prompt} & \textbf{Label Names} \\
\midrule
\textbf{RTE}
    & His family has steadfastly denied the charges. \newline\vspace{2.6pt}\hspace{-0.1cm} question: The charges were denied by his family. True or False? \newline\vspace{2.6pt}\hspace{-0.1cm} answer: \underline{True} &  True, False \\

\midrule
    & The charges were denied by his family? \newline\vspace{2.6pt}\hspace{-0.1cm} His family has steadfastly denied the charges. \newline\vspace{2.6pt}\hspace{-0.1cm} Therefore, the answer is \underline{yes}. &  yes, no \\

\midrule
    & ``The charges were denied by his family''? \newline\vspace{2.6pt}\hspace{-0.1cm} ``His family has steadfastly denied the charges.'', so the answer is \underline{yes}. &  yes, no \\
    
\midrule
\textbf{CB}
    & He'd gone. Philip had to get them back. His Dad would kill him if he found that he'd taken them. \newline\vspace{2.6pt}\hspace{-0.1cm} question: Philip had taken them. true, false, or neither? \newline\vspace{2.6pt}\hspace{-0.1cm} answer: \underline{true} &  true, false, \newline neither \\

\midrule
    & Philip had taken them? \newline\vspace{2.6pt}\hspace{-0.1cm} He'd gone. Philip had to get them back. His Dad would kill him if he found that he'd taken them. \newline\vspace{2.6pt}\hspace{-0.1cm} Therefore, the answer is \underline{yes}. &  yes, no, \newline maybe \\

\midrule
    & ``Philip had taken them''? \newline\vspace{2.6pt}\hspace{-0.1cm} ``He'd gone. Philip had to get them back. His Dad would kill him if he found that he'd taken them.'' \newline\vspace{2.6pt}\hspace{-0.1cm} Therefore, the answer is \underline{yes}. &  yes, no, \newline maybe \\

\midrule
\textbf{WiC}
    & Room and board.
    \newline\vspace{2.6pt}\hspace{-0.1cm}
    He nailed boards across the windows.
    \newline\vspace{2.6pt}\hspace{-0.1cm}
    question: Is the word `board' used in the same way in the two sentences above?
    \newline\vspace{2.6pt}\hspace{-0.1cm} 
    answer: \underline{no}
    & no, yes \\

\midrule
    & ``Room and board.'' / ``He nailed boards across the windows.''. Similar sense of ``board''? \underline{No}.
    & No, Yes \\

\midrule
    & Room and board. He nailed boards across the windows. Does ``board'' have the same meaning in both sentences? \underline{No}.
    & No, Yes \\

\midrule
    & board.
    \newline\vspace{2.6pt}\hspace{-0.1cm}
    - ``Room and board.'' (Sense 1a)
    \newline\vspace{2.6pt}\hspace{-0.1cm}
    - ``He nailed boards across the windows.'' (Sense \underline{2a})
    & 2a, 1b \\

\bottomrule
\end{tabular}
\vspace{-0.2cm}
\caption{The different prompts we use for RTE, CB, and WiC. We \underline{underline} the token to predict. For each dataset, the first prompt is the one from GPT-3~\citep[][]{brown2020language} and the others are from~\citep{schick2020small}, modified to be compatible with left-to-right LMs when necessary.}
\label{tab:prompts}
\aboverulesep = 0.605mm
\belowrulesep = 0.984mm
\end{table*}

\subsection{SuperGLUE}
\label{ssec:SuperGLUE}

\paragraph{Datasets} Here, we go into more detail about various tasks in SuperGLUE~\citep{wang2019superglue}.
BoolQ~\citep[Boolean Questions;][]{clark-etal-2019-boolq} involves answering a yes/no question about a paragraph.
COPA~\citep[Choice of Plausible Alternatives;][]{de2019commitment} involves determining the cause (or effect) of a given premise from two possible choices.
RTE (Recognizing Textual Entailment) is a 2-sentence classification task to determine if a given premise entails a given hypothesis (2-way classification between entailed and not entailed classes)~\citep{dagan2006pascal,bar2006second,giampiccolo2007pascal,bentivogli2009fifth}.
Similarly, CB~\citep[CommitmentBank;][]{de2019commitment} is an entailment detection task but with 3 classes (entailed, contradicted, and neither).
WiC~\citep[Word-in-Context,][]{pilehvar2018wic} involves determining if a polysemous word is used with the same sense in two sentences.
WSC~\citep[Winograd Schema Challenge,][]{levesque2012winograd} is a coreference resolution task to determine the correct referrent of a pronoun in a sentence from among the provided choices.
MultiRC~\citep[Multi-Sentence Reading Comprehension,][]{khashabi2018looking} is a question-answering task where each
example consists of a context paragraph, a question about that paragraph, and a list of possible answers, multiple of which can be correct.
ReCoRD~\citep[Reading Comprehension with Commonsense Reasoning Dataset,][]{zhang2018record} is a
multiple-choice question-answering task, where each example consists of a news article and a cloze-style question about
the article in which one entity is masked out. A system must predict the masked out entity from a
list of possible entities in the provided passage.

\paragraph{Prompts Used}
In Table~\ref{tab:prompts}, we show the prompts we used for RTE, CB, and WiC in \S\ref{ssec:Is prompt selection challenging on other tasks?}.
Following \citep{schick2020exploiting}, we also vary the textual label names used to get the logits for a given output class.
I.e., for RTE, we use the logit for the word ``True'' as the probability for the ``entailed'' class and ``False'' for the ``not entailed'' class.
We compute class probabilities using a softmax over the above class logits.

\subsection{Dataset and model licenses}
\label{ssec:Dataset and Model Licenses}

LAMA is licensed under CC 4.0.\footnote{https://github.com/facebookresearch/LAMA/blob/master/LICENSE}
The licenses for SuperGLUE datasets allow for their use and redistribution in a research context (see each individual dataset papers for license details).
These datasets do not contain private, personally identifiable information but may contain offensive content.
GPT-2/DistilGPT-2 models are licensed under a modified MIT license.\footnote{https://github.com/openai/gpt-2/blob/master/LICENSE}
GPT-3 models are licensed by OpenAI API to customers via a non-exclusive, non-sublicensable, non-transferable, non-assignable, revocable license.\footnote{https://beta.openai.com/policies/terms-of-use}

\subsection{Computing MDL with ADAPET}
\label{ssec:Computing MDL with ADAPET}
For MDL as formulated in \S\ref{ssec:Minimum Description Length (MDL)}, it is not possible to evaluate on the first fold of training data, since the learning algorithm (here, finetuning) requires some initial training data.
MDL requires evaluating the loss of the learning algorithm $\mathcal{A}$ on the first fold of data without any training data.
Since finetuning is not possible without training data, we say that, in this case, $\mathcal{A}$ returns a uniform distribution over all labels, following prior work~\citep[e.g.,][]{blier2018description}.\footnote{This technique can be viewed as evaluating the labels' MDL or compression rate where the first fold is compressed using a uniform distribution rather than a learning algorithm.}
We use 16 examples (one mini-batch) in the first fold and 2 examples per fold for a remaining 8 folds, to match the number of models we train for CV.
As before, we use NLL as the loss $\mathcal{L}$, which is straightforward for most tasks.
For WSC and ReCoRD, ADAPET returns class probabilities $\in \{0, 1\}$ which we smooth as $\{\epsilon, 1-\epsilon\}$ with $\epsilon=10^{-6}$ to avoid $\infty$ loss values for CV/MDL.
For MultiRC, ADAPET makes several binary predictions per example, so we sum the NLLs for these predictions to compute per-example loss.

\subsection{Computational Cost}
\label{ssec:Computational Cost}
We use the OpenAI API to evaluate GPT-3 models, costing a total of \$2826.73 for all experiments.
For GPT-2 experiments, we use a single AMD MI50 GPU (32GB GPU memory) to perform model inference, which requires at most 8 hours (usually less) for all GPT-2/DistilGPT-2 models to evaluate $\mathbb{E}_{R,F}[\text{CV}(\mathcal{A}, R, F)]$, $\mathbb{E}_{R,F}[\text{MDL}(\mathcal{A}, R, F)]$, and expected test accuracy for LAMA and SuperGLUE (any number of training examples).
For ADAPET experiments, we use a single AMD MI50 GPU for up to 12 hours to run training and inference for a single model and hyperparamater setting.

\end{document}